\documentclass{article}

\usepackage[main, final]{neurips_2026}

\usepackage[utf8]{inputenc} 
\usepackage[T1]{fontenc}    
\usepackage{hyperref}       
\usepackage{url}            
\usepackage{booktabs}       
\usepackage{amsfonts}       
\usepackage{nicefrac}       
\usepackage{microtype}      
\usepackage{xcolor}         

\usepackage{booktabs}
\usepackage{array}
\usepackage{caption}
\usepackage[T1]{fontenc}
\usepackage{graphicx}
\usepackage{enumitem}
\usepackage{amsmath}
\usepackage{tabularx}
\usepackage{placeins}
\usepackage{makecell}
\usepackage{longtable}

\newcolumntype{C}[1]{>{\centering\arraybackslash}p{#1}}
\newcolumntype{Y}{>{\raggedright\arraybackslash}X}
\newcommand{\appendixtablestyle}{%
  \centering
  \small
  \setlength{\tabcolsep}{4pt}
  \renewcommand{\arraystretch}{1.15}
}

\newcommand{\standardtable}{}

\title{Directional Confusions Reveal Divergent Inductive Biases Through Rate-Distortion Geometry in Human and Machine Vision}

\author{%
  \begin{minipage}[t]{0.30\textwidth}
    \centering
    Leyla Roksan Caglar \\
    {\normalfont
    Windreich Department of AI \\
    and Human Health \\
    Icahn School of Medicine \\
    at Mount Sinai \\
    New York, NY, USA \\
    {\footnotesize\texttt{leylaroksan.caglar@mssm.edu}}}
  \end{minipage}%
  \hspace{0.03\textwidth}%
  \begin{minipage}[t]{0.30\textwidth}
    \centering
    Pedro A.M. Mediano \\
    {\normalfont
    Department of Computing \\
    Imperial College London \\
    London, UK \\
    {\footnotesize\texttt{p.mediano@imperial.ac.uk}}}
  \end{minipage}%
  \hspace{0.03\textwidth}%
  \begin{minipage}[t]{0.30\textwidth}
    \centering
    Baihan Lin \\
    {\normalfont
    Windreich Department of AI \\
    and Human Health \\
    Department of Psychiatry \\
    Icahn School of Medicine \\
    at Mount Sinai \\
    New York, NY, USA \\
    {\footnotesize\texttt{baihan.lin@mssm.edu}}}
  \end{minipage}
}

\begin{document}

\maketitle

\begin{abstract}
To humans, a robin seems more like a bird than a bird seems like a robin, but does this asymmetry also hold for machine vision? Humans and modern vision models can match each other in accuracy while making systematically different kinds of errors, differing not in how often they fail, but in who gets mistaken for whom. We show that these directional confusions reveal distinct inductive biases invisible to accuracy alone. Using matched human and deep neural network responses on a natural-image categorization task under 12 perturbation types, we quantify asymmetry in confusion matrices and link its organization to the geometry of the information--error trade-off — how efficiently, and how gracefully, a system generalizes under distortion. We find that humans exhibit broad but weak asymmetries across many class pairs, whereas deep vision models show sparser, stronger directional collapses into a few dominant categories. Training on single distortions or noise mixtures brings overall asymmetry into the human range without recovering the human-like distributed structure. Generative simulations further show that these two asymmetry organizations shift the trade-off geometry in opposite directions even at matched accuracy, explaining why the same scalar asymmetry score can reflect fundamentally different generalization strategies. Together, these results establish directional confusion structure as a sensitive, interpretable signature of inductive bias that accuracy-based evaluation cannot recover.
\end{abstract}

\section{Introduction}
People judge a robin to be more similar to a bird than a bird is to a robin. An ellipse seems more like a circle than a circle seems like an ellipse \citep{tversky1982similarity}. These directional asymmetries are not noise, they reflect the graded structure of cognitive representations — prototype effects \citep{rosch1975cognitive}, feature salience \citep{tversky1977features}, category typicality \citep{rosch1975family} — and they violate the assumptions of symmetric similarity spaces \citep{shepard1964attention,shepard1987toward}. Directional confusions in categorization tasks \citep{nosofsky1986attention,nosofsky1991stimulus,getty1979prediction,kahana2002recognizing} can therefore serve as diagnostic signatures of representational bias and efficiency \citep{sims2018efficient,jakob2023rate}.

Do artificial vision systems exhibit the same asymmetric structure or a fundamentally different one? Humans and modern ANNs increasingly reach similar categorical accuracy, but most evaluation pipelines treat confusion as unstructured, asking how often a system errs rather than who gets mistaken for whom and in which direction \citep{gupta2021visual,attarian2020transforming}. Yet, this kind of directional structure is precisely where inductive biases leave their fingerprint, reflecting the priors a system implicitly imposes when mapping ambiguous or degraded inputs to categories.

However, evaluations of ANNs under distribution shift typically abstract away error structure. Metrics such as accuracy or top-$k$ error implicitly treat confusion patterns as symmetric or irrelevant \citep{geirhos2020shortcut,recht2019imagenet}, even though deep vision models can produce highly structured one-way failures, such as collapsing diverse subcategories into a dominant prototype (e.g., many dog breeds into "Labrador") without the reverse confusion occurring \citep{shankar2020evaluating,miller2021accuracy,d2025geometry}. Enhanced under adversarial noise or texture-based corruptions, models may confuse texture-diagnostic categories (e.g., "zebra" with "barcode") in one direction only \citep{geirhos2019imagenet,ilyas2019adversarial}. Large-scale evaluations show that this gap between human and machine error consistency persists even in state-of-the-art models \citep{geirhos2021partial}, and simplicity bias — the tendency of ANNs to rely on the simplest predictive features available — provides a mechanistic account of why such structured failures arise \citep{shah2020pitfalls}. The most direct precedent, \citet{liu2025human}, compares human and machine confusion matrices at the level of aggregate error distributions, but does not characterize the directional organization of asymmetries or link error structure to generalization properties.

This raises a specific question: when vision systems are matched for task, perturbation, and accuracy level, do humans and ANNs exhibit systematically different directional confusion structure, and does that structure predict generalization geometry independently of accuracy? Cognitive accounts explain human asymmetries via graded similarity and attentional biases \citep{tversky1977features,rosch1975cognitive}, predicting broad but weak directional tendencies spread across many class pairs. By contrast, prototype collapse and shortcut-driven feature reliance in ANNs \citep{shankar2020evaluating,geirhos2020shortcut,shah2020pitfalls} predict sparser but stronger sink-like failures concentrated on a few classes — representations collapsed onto a small set of dominant decision boundaries rather than the graded similarity gradients that characterize human-like representations \citep{sorscher2022neural,wei2025representational}. These two organizations reflect fundamentally different priors. Distributed asymmetries indicate sensitivity to graded similarity across many feature dimensions, while sink-like collapses indicate rigid reliance on a small set of dominant decision boundaries. Critically, two systems can show similar accuracy (or even similar aggregate asymmetry) while failing for qualitatively different reasons, with different implications for downstream reliability. We therefore not only compare asymmetry structure across systems, but explicitly test, in simulation at matched accuracy, whether the organization of directional errors carries information about generalization that is independent of accuracy.

\begin{figure*}[t]
  \centering
  \includegraphics[width=\textwidth]{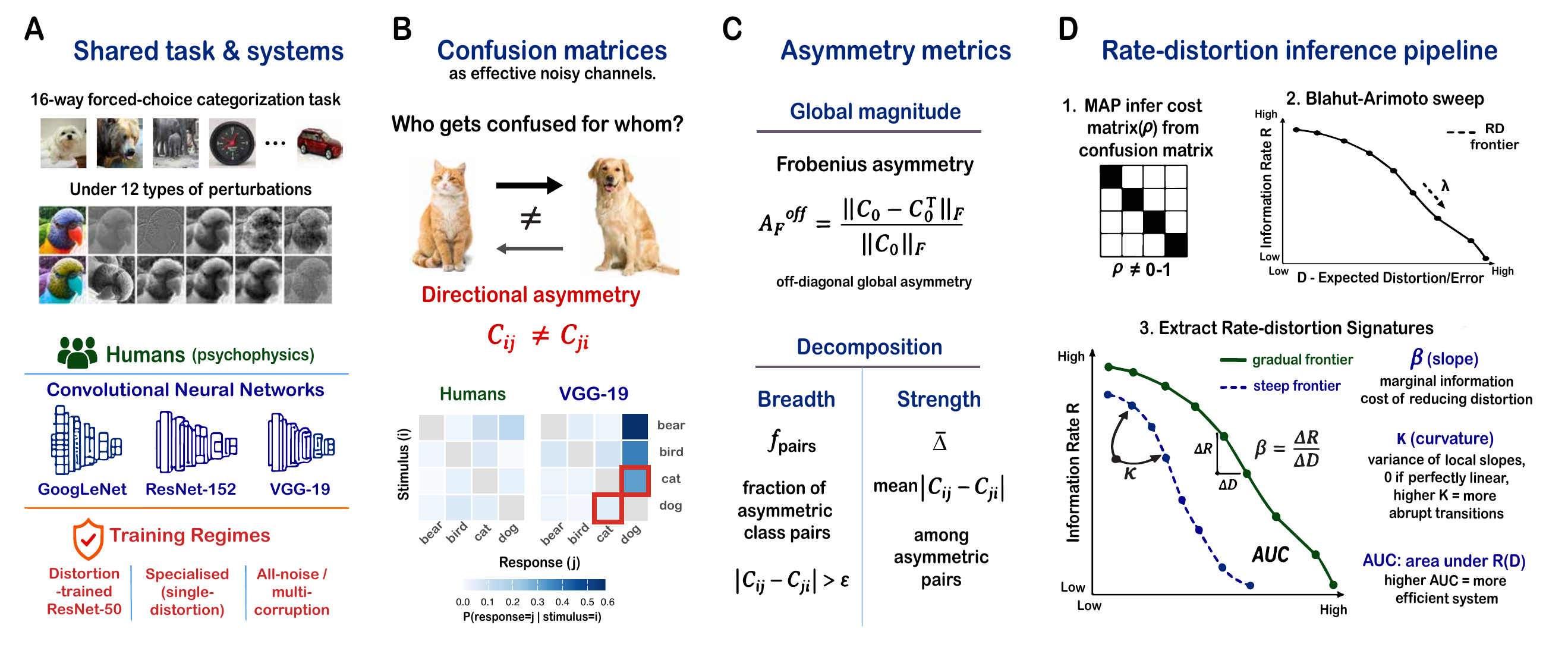}
  \caption{\textbf{Analysis pipeline.} \textbf{(A)}~Matched human and ANN stimulus--response behavior under perturbation. \textbf{(B)}~Confusion-matrix construction, illustrating the contrast between human broad--weak errors and ANN sink-like collapses (noise perturbation condition, $\sigma=0.10$). \textbf{(C)}~Schematic asymmetry decomposition into breadth and strength, alongside the global Frobenius magnitude. \textbf{(D)}~Schematic RD frontier inference yielding three geometric signatures ($\beta$, $\kappa$, AUC).}
\label{fig:schematic}
\end{figure*}

To formalize this, we adopt a rate--distortion (RD) framework that treats each system as a noisy communication channel and summarizes its generalization behavior as a geometric signature of the information--error trade-off — how efficiently, and how gracefully, a system compresses stimulus--response information under distortion (see Section~\ref{sec:rd_framework}). Within this formalism, the \emph{organization} of asymmetry, not just its magnitude, is predicted to shape where a system sits on the RD frontier, linking the structure of directional errors directly to generalization geometry. We ask whether humans and ANNs dissociate in asymmetry organization, how that structure relates to generalization geometry independently of accuracy, and whether robustness training closes the gap. We find that (i) humans distribute confusion broadly and weakly across many class pairs while ANNs concentrate failures into a few strong, sink-like collapses, a dissociation invisible to accuracy; (ii) asymmetry covaries with generalization geometry across systems, but scalar-asymmetry couplings to RD signatures are largely accounted for by accuracy, so the same performance can mask different failure regimes; (iii) generative simulations show the two organizations produce opposite effects on generalization geometry even at matched accuracy, explaining why scalar asymmetry metrics are insufficient; and (iv) robustness training reduces overall asymmetry but does not restore the human organization, falling below the human level on breadth, strength and error concentration alike.

\section{Rate-Distortion Framework}
\label{sec:rd_framework}
To formalize how directional error structure relates to generalization, we adopt a rate–distortion (RD) framework grounded in information theory and efficient coding \citep{shannon1948mathematical,sims2018efficient}. We treat each system --- human or machine --- as an effective noisy communication channel whose behavior is summarized by its stimulus–response confusion matrix. The central object is the information–error trade-off: how much mutual information between stimulus and response must be preserved to achieve a given level of categorical distortion. Tracing this trade-off yields a rate–distortion (RD) frontier \citep{shannon1948mathematical}, and the shape of that frontier characterizes how efficiently and how gracefully a system generalizes under compression \citep{sims2018efficient} (see Fig.~\ref{fig:schematic} for an illustrated inference pipeline).

Concretely, we infer a latent distortion matrix $\rho \in \mathbb{R}_{\geq 0}^{K \times K}$ from each observed confusion matrix via maximum-a-posteriori (MAP) estimation, using Blahut--Arimoto (BA) iterations \citep{blahut1972computation,arimoto1972algorithm} to trace the optimal channel at each compression level. Sweeping an inverse-temperature parameter $\lambda$ over the resulting frontier yields three compact geometric signatures that are behavioral read-outs of the inferred cost geometry:

\setlength{\parskip}{0pt}
\begin{itemize}[leftmargin=*, itemsep=1pt, topsep=2pt]
  \item \textbf{Slope} ($\beta$): the median finite-difference derivative $\Delta R / \Delta D$ along the frontier --- how steeply error falls as the system retains more stimulus--response information. Because $R$ decreases with $D$, $\beta$ is negative, and a more negative value means a steeper trade-off.
 \item \textbf{Curvature} ($\kappa$): the variance of local slopes, capturing how nonuniform the trade-off is across operating points. Low $\kappa$ corresponds to graceful degradation as information is reduced. High $\kappa$ to an abrupt, threshold-like collapse.
  \item \textbf{Efficiency} (AUC): the trapezoidal area under $R(D)$ across the swept $\lambda$ range. A larger AUC means the system retains more stimulus--response information at each level of tolerated error, so class distinctions are preserved rather than lost as error tolerance varies. Because the prior fixes the scale of $\rho$, the distortion range covered by the fixed $\lambda$ grid is comparable across systems (Appendix~\ref{app:map_prior}).
\end{itemize}

A key property of this framework is that it accommodates asymmetric confusion matrices directly, without forcing behavioral data into a symmetric similarity space. This makes it a natural formalism for the present analysis. The RD frontier is shaped by the full structure of the distortion matrix — including its asymmetric component — so systems with different asymmetry organizations are expected to trace qualitatively different frontiers, even at matched accuracy levels. The organization of directional errors, not just their magnitude, is therefore a first-class input to generalization geometry (Fig.~\ref{fig:schematic}, panels~B-D).
The analysis rests on three assumptions: (i) the behavioral channel is adequately summarized by the confusion matrix; (ii) the cost of confusing two classes is graded rather than binary, and is inferred under a structured, component-wise prior (Appendix~\ref{app:map_prior}); and (iii) the frontier is generated by efficient coding under that cost, the setting validated on human identification data by \citet{sims2018efficient}. Recovering $\rho$ from a finite confusion matrix is not guaranteed to be unique, so we do not claim to recover $\rho$ or the full frontier. Rather, the claim is that the three low-dimensional signatures are recoverable, which we test directly in simulations with known ground truth (Section~\ref{sec_sim}).

\vspace{-10pt}
\section{Methods and Modeling Framework}

\subsection{Datasets, Perturbations, and Evaluated Systems}
We analyze matched human and ANN model behavior on controlled perturbations of natural images in a $K=16$ ImageNet-derived categorization setting. The primary stimulus benchmark is the Generalization repository (GEN; \citep{geirhos2018generalisation}), which includes twelve perturbation families (e.g., colour/grayscale, contrast, filtering, phase noise, power equalisation, rotation, Eidolon variants, and uniform noise), each parameterized by distortion strength to produce systematic out-of-distribution (OOD) conditions. The data includes $\sim$83k human psychophysics trials, as well as three baseline pretrained convolutional neural networks (CNNs): GoogLeNet \citep{szegedy2015going}, ResNet-152 \citep{he2016deep}, and VGG-19 \citep{simonyan2015verydeep}. To study training-induced variation, we evaluate models with different robustness regimes. In particular, we include (a) \textit{Distortion-trained} ResNet-50 models trained from scratch with distorted training distributions, (b) \emph{Specialised} single-distortion models trained on one perturbation family and evaluated across all perturbations, and (c) \emph{All-noise / multi-corruption} models trained on mixtures of noise-like corruptions and evaluated across individual perturbations (see the GEN repository \citep{geirhos2018generalisation} for further details on benchmark stimuli, model training, and task evaluations).

\subsection{Confusion Matrices as Effective Behavioral Channels}
\label{sec:confusion_matrices}
We treat each system's confusion matrix as defining an effective noisy channel, then ask what latent distortion structure — and what information–error trade-off — is implied by its pattern of errors, including their directional asymmetries (Fig.~\ref{fig:schematic}, 
panel~B). For each system $s$, experiment $e$, and distortion level $d$, we summarize stimulus--response behavior with a $K\times K$ confusion matrix $N^{(s,e,d)}$, where $N_{ij}$ counts responses of class $j$ to stimuli of class $i$. Row-normalization yields an empirical conditional distribution
$C^{(s,e,d)}_{ij} = \Pr_s(y=j \mid x=i;\, e,d) \approx N^{(s,e,d)}_{ij} / \sum_{j'} N^{(s,e,d)}_{ij'}$,
where $C^{(s,e,d)}_0$ denotes $C^{(s,e,d)}$ with its diagonal entries set to zero (i.e., excluding correct responses). Because $C$ is built from hard (argmax) decisions, it is invariant to argmax-preserving recalibration of a model's posterior (e.g., temperature scaling), so calibration is largely orthogonal to the signatures reported here.
We infer the latent distortion matrix $\rho$ by maximum-a-posteriori (MAP) estimation, with the likelihood evaluated via the Blahut Arimoto-optimal channel \citep{blahut1972computation,arimoto1972algorithm} under $\rho$ (scale absorbed into $\rho$), following \citet{sims2018efficient} and adapted here to directional asymmetry.
To trace the rate--distortion (RD) frontier, we scale $\rho$ by an inverse-temperature parameter $\lambda>0$ and compute the corresponding optimal channel. Specifically, we solve for the RD-optimal channel $q_\lambda(y|x)$ via Blahut--Arimoto fixed-point updates: $q_\lambda(y|x) \propto p(y)\exp(-\lambda\rho(x,y))$, iterated jointly with $p(y) = \sum_x p(x)q_\lambda(y|x)$ until convergence.
We then trace the frontier over 50 log-spaced $\lambda$ values ($\lambda\in[10^{-3},10^{3}]$), computing the rate in bits, and extract the three signatures defined in Section~\ref{sec:rd_framework} (full specification in Appendix~\ref{app:map_prior}).

\subsection{Quantifying Asymmetry}
\label{sec:asym_metrics}
We define directional asymmetry as deviation from matrix symmetry in the row-normalized confusion matrix $C$, with a threshold $\varepsilon = 10^{-12}$ to suppress numerical noise (Fig.~\ref{fig:schematic}, panel~C). We use three complementary measures. \textbf{Breadth}: $f_{\mathrm{pairs}} = n_{\mathrm{pairs}}/\binom{K}{2}$, where $n_{\mathrm{pairs}} = \sum_{i<j}\mathbb{I}[|C_{ij}-C_{ji}|>\varepsilon]$, captures how many class pairs exhibit any directional asymmetry. \textbf{Strength}: $\bar{\Delta} = \mathbb{E}_{i<j}[|C_{ij}-C_{ji}|\mid |C_{ij}-C_{ji}|>\varepsilon]$ captures the mean magnitude of directional deviation among asymmetric pairs, where $\mathbb{E}_{i<j}[\cdot]$ denotes the sample mean over all class pairs $i<j$. \textbf{Global magnitude}: quantified by the normalized off-diagonal Frobenius norm $A_F^{\mathrm{off}}(C) = \|C_0 - C_0^\top\|_F / \|C_0\|_F$, which excludes correct responses from both numerator and denominator and is used as the asymmetry predictor in all analyses reported here. Implementation details and the choice of $\varepsilon$ are in Appendix~\ref{app:asym_impl}. All analyses use block-wise aggregates (experiment $\times$ condition $\times$ model instance) to avoid pseudo-replication.
We compare groups using Wilcoxon rank-sum tests for robustness to non-normality and Welch's $t$-tests for effect size estimation, with $p$-values corrected via Benjamini--Hochberg FDR (BH--FDR) and 95\% CIs via the Welch--Satterthwaite approximation.

\subsection{Linking Asymmetry to RD Geometry}
\label{sec:asym_rd}
We test whether directional confusability covaries with RD behavior \textbf{(Fig.~\ref{fig:schematic}, panel~D)}, using $A_F^{\mathrm{off}}(C)$ (Section~\ref{sec:asym_metrics}) as the asymmetry predictor for each block. Channels with near-deterministic rows are excluded on entropy and response-dominance criteria (Appendix~\ref{app:collapse_filter}). Asymmetry--RD relationships are estimated via rank correlations, block-demeaned interaction models and accuracy-controlled regressions with cluster-robust inference (Appendices~\ref{app:asym_rd_models} and~\ref{app:inference}).

\subsection{Generative Simulation Linking Asymmetric Inductive Bias to RD Signatures}
\label{sec_sim}
Our empirical results quantify asymmetry and RD geometry but do not expose the generative mechanisms behind their relationship. We therefore simulate systems with tunable asymmetric distortion structures to test interpretability and recoverability. Specifically, we predict that broad–-weak and sink-like asymmetry organizations will produce opposite effects on RD geometry. Distributed asymmetries should expand the RD frontier by preserving information across many class distinctions, while concentrated sink-like asymmetries should collapse it by funneling probability mass into a few dominant responses. This dissociation should persist even after controlling for overall accuracy.

We fix $K=16$ classes. Each replicate involves a ground-truth distortion matrix $\rho_{\mathrm{true}}$, simulated confusion counts $N$, inferred distortion $\hat\rho$, and derived metrics. We construct $\rho_{\mathrm{true}} = \rho_{\mathrm{sym}} + aA$ where
$\rho_{\mathrm{sym}}$ is symmetric, $A$ is skew-symmetric ($A=-A^\top$),
$\rho_{ii}=0$, and $a \geq 0$ controls antisymmetry magnitude. We evaluate two antisymmetry types: (i) \textit{broad--weak} (dense skew-symmetric noise), and (ii) \textit{sink-like} (targeted bias toward a small set of classes).
Given $\rho_{\mathrm{true}}$, we generate channels via Blahut--Arimoto iterations, draw confusion counts from a multinomial distribution, and recover $\hat{\rho}$ using the same MAP pipeline as for empirical systems. Simulations sweep antisymmetry magnitude $a$, generation inverse temperature $\lambda_{\mathrm{gen}}$, and per-class trial count $N_{\mathrm{per\,row}}$ across both structures with multiple seeds. Full grid details, secondary diagnostics, and sensitivity checks are reported in Appendix~\ref{sec:sim_secondary_diagnostics} (Tables~\ref{tab:sim_recovery_A3}--\ref{tab:sim_sign_stability_A6}).

\begin{figure*}[t]
  \centering
  \includegraphics[width=\textwidth]{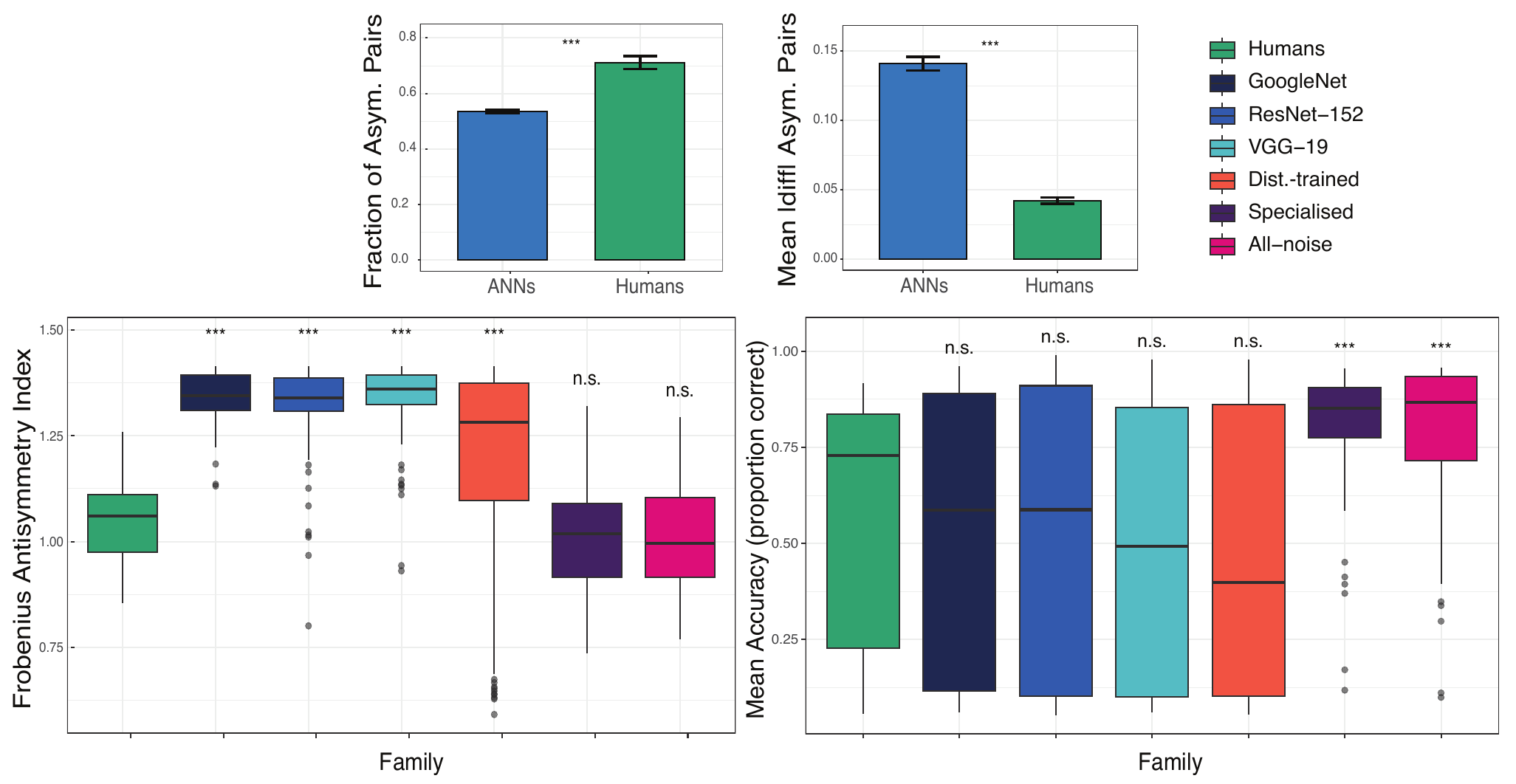}
  \caption{\textbf{Asymmetry decomposes into \textit{breadth} vs.\ \textit{strength}, revealing a dissociation between humans and ANNs invisible to accuracy.} \textbf{Top:} Breadth ($f_{\mathrm{pairs}}$, left) and strength ($\bar{\Delta}$, right) of directional structure, comparing all ANNs pooled against humans. Error bars show s.e.m.\ across blocks; significance marks two-sided Wilcoxon rank-sum tests. \textbf{Bottom:} Frobenius asymmetry index (left) and mean accuracy (right) by model family, each compared individually against humans (planned Wilcoxon tests, BH--FDR corrected). Each point is one block (experiment$\times$condition$\times$model). Groups significantly more asymmetric than humans show no accuracy difference (n.s.), and vice versa, confirming a double dissociation between asymmetry organization and accuracy.}
\label{fig:asym_decomposition}
\end{figure*}

\section{Results}

\subsection{Asymmetry magnitude and sparsity dissociate between humans and ANNs.}
\label{sec:results_asym}
We quantified asymmetry using block-wise summaries (one value per unique \emph{experiment}$\times$\emph{condition}$\times$\emph{model}; ANNs: $n=1569$ blocks; humans: $n=81$ blocks; representative matrices in Fig.~\ref{fig:confusion_asymmetry_K16}). ANNs exhibited larger global asymmetry than humans on the Frobenius index, yet sparser directional structure (all comparisons Wilcoxon rank-sum, $p\le6.6\times10^{-5}$; block counts and group means in Table~\ref{tab:asymmetry_appendix_summary_A1}, per-group test statistics in Table~\ref{tab:group_tests_A7}, and \textbf{Fig.~\ref{fig:asym_decomposition}}). Humans spread directionality across more class pairs than ANNs (fraction of asymmetric pairs $0.712$ vs.\ $0.535$, a gap that widens against baseline CNNs alone), whereas conditional on a pair being asymmetric ANNs showed over three times the per-pair deviation ($\bar{\Delta}=0.141$ vs.\ $0.0422$). Global asymmetry magnitude therefore conflates two separable properties on which humans and ANNs dissociate in opposite directions: breadth (more pairs in humans) and strength (larger deviations in ANNs; \textbf{Fig.~\ref{fig:schematic}, panel C}). 

Planned humans vs.\ model comparisons of Frobenius asymmetry (BH--FDR across groups) indicated that baseline CNNs and the Distortion-trained regimes remain significantly more asymmetric than humans, whereas the specialised and all-noise regimes were not significantly different from humans \textbf{(Fig.~\ref{fig:asym_decomposition})}, suggesting that robustness-oriented training can reduce global asymmetry toward the human range (Table~\ref{tab:group_tests_A7}). That reduction does not, however, recover the human breadth--strength organization (Section~\ref{sec:results_robust}). Importantly, the dissociation runs in both directions (\textbf{Fig.~\ref{fig:asym_decomposition}}, bottom-right). Groups that are significantly more asymmetric than humans (GoogLeNet, ResNet-152, VGG-19, Distortion trained) show no significant difference from humans in accuracy, whereas groups that match humans on asymmetry (Specialised, All-noise) are significantly more accurate than humans. This double dissociation shows that group differences in asymmetry are not simply group differences in accuracy, and that directional confusion structure carries information not captured by accuracy-based evaluation.

To assess generalizability beyond classic CNNs, we replicated the Frobenius asymmetry analysis on 15 modern architectures including Vision Transformers, large CNNs, and self/semi-supervised models. All families remained significantly more asymmetric than humans (all p < 0.001, BH-FDR; \textbf{Fig.~\ref{fig:asym_modelzoo}}, see Appendix~\ref{sec:architectures_appx}) regardless of accuracy level, confirming the dissociation is not architecture-specific (per-condition matrices; the full breadth--strength analysis on pooled matrices is in Section~\ref{sec:results_robust}).

\begin{figure*}[t]
  \centering
  \includegraphics[width=\textwidth]{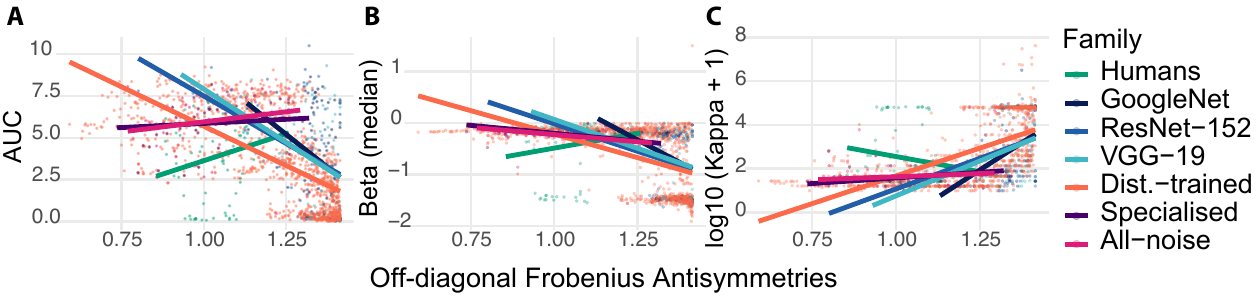}
 \caption{\textbf{Directional confusion asymmetry covaries with RD signatures across humans and model families.} These pooled couplings are largely accounted for by accuracy. Each point is one block (experiment$\times$condition$\times$model) in the GEN corpus ($n=1610$); $x$-axis is off-diagonal Frobenius asymmetry $A_F^{\mathrm{off}}$. Thick lines show family-wise linear trends. Raw couplings are shown here; the accuracy-controlled versions, and the same analysis in the ModelZoo corpus, are in Fig.~\ref{fig:asym_rd_before_after}.\textbf{(A)}~Efficiency (AUC). Greater asymmetry coincides with reduced RD efficiency in most ANN families, with a distinct human profile. \textbf{(B)}~Slope ($\beta$). Directional confusions track steepness of the information--error trade-off. \textbf{(C)}~Curvature ($\log_{10}(\kappa+1)$). Regime-dependent coupling between directional structure and RD nonlinearity. All $\kappa$ values are strictly positive (range: 8.6--$3.5\times10^9$). Apparent sub-zero region of the $y$-axis reflects extrapolation of group-wise linear trends beyond the data range.}
 \label{fig:asym_decomposition_2}
\end{figure*}

\subsection{Robustness of the dissociation}
\label{sec:results_robust}

\paragraph{Robustness training lowers strength without restoring breadth.}
The two regimes that match humans on global asymmetry reach that match by a different route. In the GEN corpus the specialised and all-noise regimes reach conditional strengths of $0.030$ and $0.025$, at or below the human value of $0.042$, but their breadth reaches only $0.503$ and $0.583$ against $0.712$ for humans --- closer to the baseline CNNs ($0.446$--$0.457$). Robustness training therefore produces a third organization, narrow and weak, rather than the broad and weak human one, and it is this combination that brings the scalar Frobenius index into the human range (Table~\ref{tab:group_tests_A7}). Matching on a one-number asymmetry score does not imply matching on its organization.

\paragraph{ANN errors funnel into several shifting receivers, not a single sink.}
A natural alternative is that ANN sparsity reflects a single ``sink'' category. It does not: error mass above chance falls on 4.95 classes on average (humans: 6.07), and the dominant receiver shifts across perturbations (modal-receiver consistency 0.21--0.46 across models, versus 0.73 for humans), while ANNs nonetheless concentrate more mass into their most-confused class (top-1 excess over a density-matched null: 0.281 vs.\ 0.086; Appendix~\ref{app:sink}). The organization is therefore a set of strong, shifting directional collapses rather than a fixed bin, and it persists under accuracy control and in robustness-training.

\paragraph{Concentrated error structure persists under density-matched comparison.}
To test whether ANN error concentration is a property of their errors rather than of the $K=16$ benchmark, we evaluated three architectures (ResNet-50, ViT-B/16, and a self-supervised ConvNeXt-V2) on the full ImageNet-1k validation set under three ImageNet-C corruption families at two severities --- a $60\times$ larger label space, on different images and categories (also see Appendix~\ref{app:label_grouping} for robustness to how the original 16 labels are grouped). Against references generated from our own broad--weak and sink cost geometries at matched per-class error and trial counts, real ANN errors concentrate onto ${\sim}4$--$24$ effective target classes versus ${\sim}46$ for the density-matched diffuse reference, in all 11 conditions with adequate error density (Table~\ref{tab:imagenet1k_conditions}, Fig.~\ref{fig:concentration_K1000}). The structure is concentrated but not degenerate, falling well short of the 1--2 effective targets of the extreme sink reference (Appendix~\ref{app:imagenet1k}).

\paragraph{Modern architectures do not close the gap.}
Per-condition ModelZoo matrices are too sparse for the breadth--strength decomposition, so we pooled each perturbation family across its severity levels and repeated the analysis on 15 modern systems. Every architecture family was significantly less broad, higher in strength, more Frobenius-asymmetric and sparser than humans, with all 24 family$\times$metric comparisons surviving BH--FDR correction. Vision Transformers and semi-supervised models are the least broad families (0.469 and 0.451). Critically, the systems this benchmark ranks as most human-like on error consistency (CLIP, BiT-M, ViT-L, SWAG, ResNeXt-SWSL, EffNet-L2) are not human-like in asymmetry organization: breadth 0.495 vs.\ 0.923 and Frobenius 1.302 vs.\ 0.882 against humans (both $p_{\mathrm{BH}}\le10^{-7}$). Directional structure therefore carries inductive-bias information that accuracy and error-consistency benchmarks do not (Table~\ref{tab:modelzoo_families}, Fig.~\ref{fig:modelzoo_breadth_strength}).

\paragraph{The human profile holds at the individual-observer level.}
Every individual observer's mean strength (0.024--0.043) falls entirely below the baseline CNN and distortion-trained range in the same corpus (0.121--0.154), so the low-strength half of the human profile is not an artifact of pooling observers. Individual breadth is lower than the pooled value, but a trial-count control indicates this is a density effect rather than a strategy difference (Appendix~\ref{app:per_observer}). Cross-observer correlations of per-experiment breadth are uniformly high (0.90--0.99), with no evidence of distinct strategy subgroups.

\subsection{Asymmetry--RD coupling and its dependence on accuracy.}
\label{sec:results_rd}
Asymmetry organization related systematically to RD trade-off geometry, but the strength and direction of coupling depended on model family and training regime. Because naive correlations partly reflect shared dependence on accuracy, we report both pooled associations and accuracy-controlled within-block regressions (specification in Appendix~\ref{app:asym_rd_models}; cluster-robust inference and collapsed-channel exclusion in Appendix~\ref{app:inference}). In the empirical data, pooled asymmetry--RD associations are largely accounted for by accuracy, signature by signature below. Because power differs across groups ($n=1182$ Distortion-trained blocks versus $n\le81$ elsewhere), a non-significant residual is not evidence of absence, and the accuracy-independent evidence therefore rests mainly on the generative simulations (Section~\ref{sec_sim}), which isolate asymmetry organization at matched accuracy.

\paragraph{Efficiency (AUC).}
Greater asymmetry was associated with lower RD efficiency across most ANN families, but this coupling was largely accounted for by shared dependence on accuracy --- with one exception \textbf{(Fig.~\ref{fig:asym_decomposition_2}A)}. Pooled rank correlations were strongly negative for Distortion-trained models (\(\rho=-0.73\), \(n=1182\)) and negative for all three baseline CNNs ($\rho=-0.39$ to $-0.43$), and block-demeaned interaction models showed substantially steeper within-block dependence for the Distortion-trained regime than for humans (\(\Delta\mathrm{slope}=-7.89\), \(p=1.27\times10^{-7}\)), with the same direction for Specialised and All-noise (Table~\ref{tab:asym_rd_by_group_A2}). After accuracy control, however, only the Distortion-trained regime retains a residual association, and its sign reverses: the conditional association is \emph{positive} (slope \(=1.02\pm0.30\), \(t=3.45\), \(p_{\mathrm{BH}}=0.010\)), while humans and baseline models show no reliable accuracy-controlled slopes (all $p_{\mathrm{BH}}\ge 0.147$). This is the only association significant under every inference scheme and $\lambda$ grid we tested. The negative pooled associations are therefore largely an accuracy effect, and the same pattern holds in both corpora (Fig.~\ref{fig:asym_rd_before_after}).

\paragraph{Slope (\(\beta\)).}
A similar pattern held for RD slope: asymmetry and $\beta$ were negatively correlated across most regimes, (Distortion-trained $\rho=-0.57$; baseline CNNs $-0.33$ to $-0.46$), with steeper within-block dependence for the robustness-trained regimes than for humans (Distortion-trained $\Delta\mathrm{slope}=-1.21$, $p=0.0085$), but these associations were largely accounted for by accuracy (Table~\ref{tab:asym_rd_by_group_A2}; \textbf{Fig.~\ref{fig:asym_decomposition_2}B}). Under accuracy control, Distortion-trained models no longer showed a reliable asymmetry--$\beta$ relationship (slope \(=0.07\), \(p_{\mathrm{BH}}=0.77\)), and baseline CNNs showed none (all \(p_{\mathrm{BH}}\ge 0.36\)). Specialised models showed a negative association (slope \(=-0.79\pm0.26\), \(t=-3.08\), \(p_{\mathrm{BH}}=0.037\)), but this is the one result in our analysis that is not robust to the $\lambda$ grid: under a narrower grid it attenuates to approximately $-0.13$ and is no longer significant (Appendix~\ref{app:map_prior}). We therefore do not interpret it as an accuracy-independent effect.

\paragraph{Curvature (\(\kappa\)).}
Curvature showed the most accuracy-dependent pattern of the three signatures. The Distortion-trained regime showed a strong positive rank association ($\rho=0.72$) and significantly steeper within-block dependence than humans ($\Delta\mathrm{slope}=3.96$, $p=1.97\times10^{-5}$), as did All-noise and Specialised (Table~\ref{tab:asym_rd_by_group_A2}), despite Specialised exhibiting an opposite pooled rank tendency \textbf{(Fig.~\ref{fig:asym_decomposition_2}C)}. None survived accuracy control: accuracy was strongly predictive of curvature (block-demeaned accuracy term $A_{\mathrm{dm}}=-4.32\pm0.41$, $t=-10.63$, $p=4.9\times10^{-11}$), whereas no group retained a significant accuracy-controlled asymmetry--curvature association (all $p_{\mathrm{BH}}\ge 0.098$; Table~\ref{tab:acc_controlled_gen}), consistent with curvature tracking performance more than directional structure.

\subsection{Generative Simulation}
\begin{figure*}[t]
  \centering
  \includegraphics[width=\textwidth]{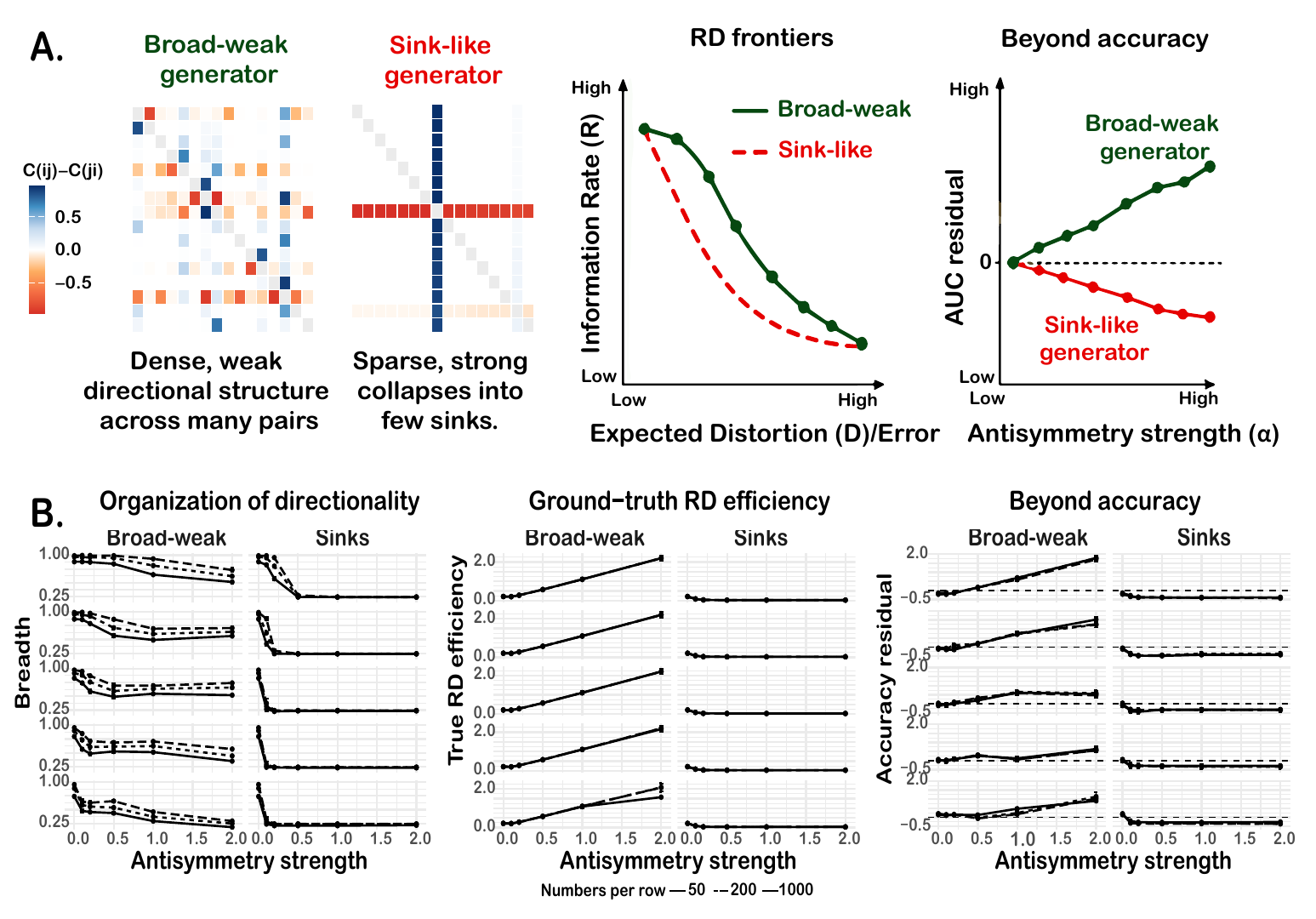}
  \caption{\textbf{Generative simulation: asymmetry organization controls RD
  geometry.} \textbf{A.} Left: simulated asymmetry matrices from one representative replicate per generator ($a=1.0$, $\lambda_{\mathrm{gen}}=2.0$,$N=200$). Right: schematic illustrations of the predicted RD frontier divergence and accuracy-residualized AUC, consistent with the simulation results in panel B. \textbf{B.} Full simulation grid. Columns compare generators; rows show five generalization regimes ($\lambda_{\mathrm{gen}}\in\{0.2,0.5,1,2,5\}$, top to bottom).}
  \label{fig:sim_mechanism_summary}
\end{figure*}

We implemented the mechanistic simulation to ask the following targeted question: \emph{When directional confusions increase, does the underlying rate--distortion (RD) geometry expand in the same way for different forms of asymmetry?} We compared two generators matched on the same control parameters (generalization regime and sample size) but differing in how directionality is organized \textbf{(Fig.~\ref{fig:sim_mechanism_summary}A)}. There was one \emph{broad--weak} mechanism that distributes weak one-way biases across many class pairs, versus a second \emph{sink-like} mechanism that concentrates probability mass into a small set of strong one-way errors.

\paragraph{Directional structure is recoverable but identifiability is mechanism-dependent.}
Across the full simulation grid (\(n=1800\) runs), numerical collapse was rare (\(138/1800=7.7\%\)), leaving \(n=1662\) non-collapsed runs for primary analyses. A stricter reliability screen, requiring that the recovered symmetric component of $\rho$ aligns with ground truth (correlation \(>0.2\)), removed an additional \(509/1662=30.6\%\), yielding \(n=1153\) strictly-recovered runs (Table~\ref{tab:sim_recovery_A3}). Recovery was strongly \emph{mechanism-dependent}, with broad--weak structure exceeding sinks in pass rate in every \((\lambda_{\mathrm{gen}},N_{\mathrm{per\,row}})\) slice and FDR significance in \(10/15\) slices (BH--FDR; Table~\ref{tab:sim_recovery_A3}). The largest gaps occurred at moderate-to-high generalization sharpness and intermediate-to-large sample sizes (maximum pass-rate gap \(0.417\); e.g.\ \(0.867\) vs \(0.450\)), indicating that sink-like directional structure is intrinsically harder to identify under the same MAP pipeline. Consequently, we report all main trends on the non-collapsed set and use strict-recovery only as a sensitivity check (Tables~\ref{tab:sim_recovery_A3} and~\ref{tab:sim_sign_stability_A6}). This differential recoverability implies that our MAP pipeline may underestimate the prevalence of sink-like structure in empirical data. Results involving sink-like regimes should therefore be interpreted with this asymmetry in mind.

\paragraph{Broad--weak versus sinks produce opposite couplings between antisymmetry and RD geometry.}
Consistent with our prediction, the same increase in antisymmetry strength produces opposite changes in RD geometry depending on whether directionality is distributed broadly or concentrated into sinks. In the broad--weak generator, increasing antisymmetry systematically \emph{expands} ground-truth RD geometry, whereas in the sink-like generator it \emph{collapses} ground-truth RD geometry toward a near-degenerate regime. This qualitative dissociation is visible in the true RD efficiency curves across regimes and remains consistent under strict-recovery filtering \textbf{(Fig.~\ref{fig:sim_mechanism_summary}; Table~\ref{tab:sim_sign_stability_A6})}. This provides a concrete generative explanation for the empirical dissociation in which humans and ANNs can exhibit comparable global asymmetry magnitude yet occupy different regions of RD space.

\paragraph{The RD--asymmetry coupling is not reducible to overall performance.}
A natural concern is that RD geometry might simply track overall performance. However, removing accuracy within each $(\lambda_{\mathrm{gen}}, N_{\mathrm{per\,row}})$ slice, the dissociation persisted: broad--weak generators show strongly positive accuracy-adjusted AUC slopes across all 15 slices (all $p_{\mathrm{FDR}}\le 7.9\times10^{-31}$), while sink slopes are negative in all 15 slices (Fig.~\ref{fig:sim_mechanism_summary}; Appendix~\ref{sec:sim_secondary_diagnostics}).

\paragraph{The breadth--strength decomposition is predictive.}
To connect mechanism to observable summaries, we decomposed directional structure into \emph{breadth} (how many class pairs exhibit directionality) and \emph{strength} (how large the directional deviation is among asymmetric pairs). After residualizing outcomes for accuracy \emph{within} \((\lambda_{\mathrm{gen}},N_{\mathrm{per\,row}})\) slices, we asked which aspect of asymmetry explains residual variation in RD signatures. For residual RD \emph{efficiency} (AUC) the component model is strong and slice-consistent: breadth and strength are both typically negative (significant in \(11/15\) and \(12/15\) slices, BH--FDR) and their interaction typically positive (\(5/15\)). Residual AUC therefore depends on how directionality is distributed --- broad-and-weak versus sparse-and-strong mixtures --- and not merely on how much of it there is. A one-number global magnitude model is by contrast nearly null for AUC residuals, reaching significance in only \(3/15\) slices. Global $A_F^{\mathrm{off}}$ thus obscures the organization that AUC is sensitive to, whereas slope and curvature track global magnitude in simulation (coefficients, ranges and per-slice counts in Table~\ref{tab:sim_residual_regressions_A4}).

\subsection{Ablations: symmetrizing a confusion matrix is not a neutral operation.}
\label{sec:ablation_results}
Confusion and similarity matrices are routinely symmetrized before analysis. We asked how much of the inferred RD geometry depends on the directional information this discards, applying two ablations --- symmetrization, and a within-row shuffle that scrambles directional targets while holding accuracy fixed --- to matched matrices from humans, a baseline CNN (ResNet-152) and a specialised model (Appendix~\ref{app:ablation}). The dependence is substantial, and its size orders the three systems exactly as their error concentration does (Fig.~\ref{fig:ablation_signatures}). Symmetrization raised $\log_{10}\kappa$ in ${\sim}90\%$ of matrices and lowered AUC in every system (Appendix~\ref{app:ablation}). Fragility, the per-matrix increase in $\log_{10}\kappa$, gives specialised (0.16) $<$ humans (0.24) $<$ ResNet-152 (0.59), with all pairwise Wilcoxon comparisons significant ($p=1.7\times10^{-6}$ to $0.029$) --- the same ordering as top-1 error concentration (0.148, 0.178, 0.412), and as the disagreement between the two arbitrary mirror directions (0.07, 0.16, 0.22), which measures the directional information a symmetry assumption discards. The shuffle sharpens the split by moving ResNet-152's curvature by $+0.35$ ($p=1.1\times10^{-5}$) while changing humans and the specialised model only slightly ($-0.06$ in both). This is not a magnitude effect, as change in $A_F^{\mathrm{off}}$ explains under $2\%$ of the variance in curvature change across all 405 ablated matrices. Removing directional structure therefore perturbs frontier curvature most where directional mass is most concentrated, so symmetrization is not a neutral preprocessing step.

\vspace{-4pt}
\section{Discussion}
Human and ANN vision systems can achieve similar categorization accuracy while relying on fundamentally different inductive biases, differing not in how often they err, but in how failures are organized. The breadth--strength dissociation we document reflects distinct priors about which features and prototypes are privileged under uncertainty. Humans distribute errors broadly, consistent with graded similarity across many feature dimensions, while ANNs concentrate failures into a small number of dominant collapse directions. Crucially, this difference is invisible to global asymmetry magnitude. Robustness-oriented training can match scalar asymmetry metrics without producing the human organization. Because it suppresses directionality across the board (i.e., below the human level) two systems can appear aligned on a scalar metric while arriving there by entirely different routes.
Treating confusion matrices as effective noisy channels and linking their directional structure to RD frontier geometry suggests why: in our simulations, breadth--strength organization specifically predicts how efficiently a system compresses stimulus--response information (AUC). Slope ($\beta$) and curvature ($\kappa$) by contrast primarily track the global magnitude of asymmetry rather than its organizational structure (Section~\ref{sec:sim_breadth_strength}). The same global asymmetry score can thus index fundamentally different generalization regimes depending on whether errors are distributed or concentrated — a distinction that accuracy-based evaluation cannot recover.

At the representational level, the breadth--strength dissociation we 
observe behaviorally would be consistent with differences in the geometry of learned feature spaces. Broad--weak asymmetries with manifolds in which categories lie along graded similarity gradients, so that many class pairs are weakly but meaningfully separated, and sink-like asymmetries with representations that collapse many inputs onto a few dominant attractor states. Establishing this link empirically is an important open direction. RD efficiency (AUC) has been shown to track penultimate-layer representational structure beyond accuracy \citep{caglar2026ratedistortionsignaturesgeneralizationinformation}, suggesting that the organization documented here may be probed without direct access to activations.

This has a direct methodological implication. Symmetric similarity and dissimilarity structure is assumed by construction across much of the field --- in representational similarity analysis, in multidimensional scaling, and wherever a confusion matrix is averaged with its transpose before analysis --- and the cognitive literature has long noted that human similarity judgments violate it \citep{tversky1977features,shepard1964attention}. Our ablations show that this assumption is not a neutral preprocessing step for the quantities studied here. Instead, symmetrizing a confusion matrix measurably changes the shape of its inferred RD frontier, and does so unequally across systems, especially for those whose errors are most concentrated. A symmetric analysis therefore discards information that is both real and differentially distributed, which can make systems with very different directional organization appear more alike than they are. Curvature therefore indexes how gracefully a system degrades, and fragility to symmetrization tracks error concentration rather than error rate.

We distinguish two kinds of conclusion. The accuracy-independent conclusions are (i) the mechanistic dissociation between asymmetry organization and RD geometry, established in simulation at matched accuracy, and (ii) the link between AUC and internal representational geometry reported in prior work \citep{caglar2026ratedistortionsignaturesgeneralizationinformation}. The conclusions that depend on accuracy are the empirical couplings of scalar asymmetry to $\kappa$ and, in large part, to $\beta$ and AUC. These are largely accounted for by accuracy, so naive correlations can partially reflect shared dependence on performance rather than structure alone. Specifically, the generative simulations are what separate organization from magnitude, formalizing a link between the organization of directional confusions and the capacity--generalization trade-offs that shape behavior. The same increase in directional asymmetry can produce opposite RD behaviors depending on how it is organized. In our simulations, broad--weak asymmetries shift the RD frontier toward higher efficiency, whereas sink-like asymmetries shift it toward lower efficiency, and these effects persist under accuracy control.

On a practical level, matching human accuracy or aggregate asymmetry is not sufficient for a human-like error organization, as systems can concentrate failures into sink-like collapses while appearing aligned on scalar metrics. Breadth and strength can serve directly as an evaluation metric computable from any confusion matrix, but using them as a training objective would additionally require a differentiable surrogate, since confusion counts rest on argmax decisions.

\vspace{-6pt}
\paragraph{Limitations and future directions.}
We note three scope conditions. First, no matched human psychophysics data exists for label spaces larger than $K=16$, so the human--model contrast is untested beyond this granularity, and at large label spaces the full directional decomposition remains under-determined at achievable error densities (Appendix~\ref{app:imagenet1k}). Second, human estimates rest on $n=81$ blocks against $n=1569$ ANN blocks, and individual-observer breadth is trial-limited (Appendix~\ref{app:per_observer}). Third, the MAP inference of $\rho$ is validated against ground truth only in simulation, where recovery is mechanism-dependent: sink-like structure is harder to identify than broad--weak structure, so our pipeline may understate its prevalence. Prior-rate sensitivity is bounded only indirectly (Appendix~\ref{app:map_prior}). Natural extensions include class-level asymmetry analysis, adversarially-trained and human-similarity-supervised models, and other task domains.

\begin{ack}
This work was supported in part through the Minerva computational and data resources and staff expertise provided by Scientific Computing and Data at the Icahn School of Medicine at Mount Sinai and supported by the Clinical and Translational Science Awards (CTSA) grant UL1TR004419 from the National Center for Advancing Translational Sciences. B.L. is supported by the Brain \& Behavior Research Foundation, the Sidney R. Baer, Jr. Foundation, the U.S. National Institutes of Health (NIH), the U.S. Department of Veterans Affairs (VA), the Hasso Plattner Foundation, and the Windreich Family Foundation.
The authors declare no competing interests.
\end{ack}

\bibliographystyle{plainnat}
\bibliography{asymmetry_bibliography}


\clearpage
\appendix


\renewcommand{\thesection}{A\arabic{section}}
\setcounter{section}{0}

\section{Compute Resources and Data Availability}
\label{app:compute_licenses}

\paragraph{Computational resources.}
The pipeline has three steps with distinct cost profiles. Step~1 --- MAP inference of the latent distortion matrix $\rho$ --- is the computational bottleneck, scaling as $O(K^2)$ in both parameter space and the inner Blahut--Arimoto loop. For $K=16$ categories, MAP inference takes approximately 50 minutes per confusion matrix on 2 CPU cores; the full empirical analysis covers 1,650 confusion matrices (all models and conditions in the GEN benchmark repository), corresponding to approximately 1,375 core-hours in total. Steps~2 and~3 --- RD frontier tracing and signature extraction ($\beta$, $\kappa$, AUC) --- run in seconds given $\rho$ and add negligible overhead. The generative simulation grid (1,800 runs across two asymmetry structures) requires approximately 10--20 CPU-hours. All jobs were parallelized across an HPC cluster, substantially reducing wall-clock time. The ImageNet-1k evaluation (Appendix~\ref{app:imagenet1k}) is the only component requiring GPU resources: three pretrained models were evaluated on the 50,000-image ImageNet validation set under six corruption conditions each, for 18 inference runs totalling 900,000 forward passes and approximately 1 GPU-hour in total on a single GPU node of the same HPC cluster. No model training was performed in this work.

\paragraph{Data and model licenses.}
The GEN benchmark~\citep{geirhos2018generalisation} is available at \url{https://github.com/rgeirhos/generalisation-humans-DNNs} under CC BY 4.0. The ModelZoo benchmark~\citep{geirhos2021partial} is available at \url{https://github.com/bethgelab/model-vs-human}. Its code is released under the MIT License, its image datasets derive from ImageNet and remain subject to the ImageNet license, and the human responses it distributes were collected as reported in \citet{geirhos2018generalisation}. All pre-trained models (GoogLeNet, ResNet-152, VGG-19, the robustness-trained variants, and the modern architectures of Appendix~\ref{sec:architectures_appx}) were accessed through ModelZoo and remain under their original licenses, listed in that repository's \texttt{MODEL\_LICENSES} file. The SWSL ResNeXt-101 weights in particular are released under CC BY-NC 4.0. MAP inference follows the pipeline of \citet{sims2018efficient}, available at \url{https://osf.io/x5ckn/}. The ImageNet-1k evaluation additionally uses the ImageNet validation set under the ImageNet terms of access, the ImageNet-C corruption benchmark~\citep{hendrycks2019benchmarking} under CC BY 4.0, and pretrained weights obtained through \texttt{timm} under the Apache License 2.0. No new models are introduced in this work and no new human data were collected.

\paragraph{Code and data availability.}
All analysis code is available at \url{https://github.com/leylaroksan/rdt_asymmetry}, including the rate--distortion inference pipeline, the asymmetry metrics, the directional-structure ablations, the generative simulation, and scripts reproducing every figure and table in this paper. The MAP inference and Blahut--Arimoto optimization were originally developed by \citet{sims2018efficient}. The RD signature extraction follows the implementation in \citet{caglar2026ratedistortionsignaturesgeneralizationinformation}, available at \url{https://github.com/leylaroksan/rdt_geometry}.
Derived data sufficient to reproduce every figure and table without re-running MAP inference --- fitted distortion matrices, RD signature summaries, collapse flags, the simulation grid outputs, and the $K=1000$ confusion matrices --- are deposited at \url{https://doi.org/10.6084/m9.figshare.34059993}. The underlying behavioral and model data remain available from the GEN and ModelZoo repositories.

\section{Statistical and Analytical Procedures}

\subsection{Collapsed Channel Filtering}
\label{app:collapse_filter}
We exclude confusion matrices with collapsed responses based on diagnostics computed from the row-normalized conditional distribution $C(y|x)$ (i.e., each row of the confusion matrix normalized to sum to 1):
\begin{itemize}
  \item Mean row entropy $< 10^{-3}$, or
  \item Mean row-maximum probability $> 0.999$.
\end{itemize}
Only a small subset of system--condition matrices were excluded, but the exclusion is consequential for $\beta$, the signature most sensitive to near-degenerate channels (the affected estimates are reported in Appendix~\ref{app:inference}).

\subsection{Asymmetry--RD Regression Models}
\label{app:asym_rd_models}
We used two linear modeling approaches to assess how asymmetry covaries with RD signatures. First, models with fixed effects per (experiment, condition) block and group-specific slopes. Second, within-block demeaning of predictor and outcome variables, followed by regression with interaction terms. Concretely, asymmetry--RD relationships are estimated via:
\begin{itemize}[leftmargin=*]
\item \textit{(i) Rank correlations:} Spearman correlations within group between $A_{\mathrm{F}}^{\mathrm{off}}$ and each RD signature.
  \item \textit{(ii) Block-controlled models:} Linear models with experiment/condition fixed effects and group-specific slopes.
  \item \textit{(iii) Accuracy-controlled models:} Because asymmetry and RD geometry both covary with overall accuracy, we additionally test whether asymmetry--RD associations persist after controlling for accuracy within matched (experiment, condition) blocks, isolating the contribution of directional structure independent of performance level.
\end{itemize}

\subsection{MAP Prior and $\lambda$ Grid}
\label{app:map_prior}
The latent distortion matrix $\rho$ is inferred by maximum-a-posteriori estimation under a structured, component-wise prior. We decompose $\rho$ into its symmetric off-diagonal part, its diagonal, and its asymmetric part, and place an independent exponential prior on each, with rates $0.1$, $10$ and $10$ respectively (equivalently, prior means $10$, $0.1$ and $0.1$). The prior is deliberately asymmetric across components: the large prior mean on the symmetric off-diagonal component reflects that confusions between distinct classes are costly, while the small means on the diagonal and asymmetric components shrink toward zero self-distortion and toward symmetry, so that directional structure must be supported by the data rather than assumed by the prior. These shared hyperparameters place all inferred $\rho$ matrices on a common scale, so that slope and distortion range are comparable across systems without rescaling $\rho$ after fitting.

Optimization is carried out on the log scale, and the posterior is approximated locally by a Laplace expansion around the MAP estimate using the numerical Hessian; the resulting mean and covariance are retained with each fit. After optimization the minimum entry of $\rho$ is subtracted to enforce non-negativity ($\rho \leftarrow \rho - \min(\rho)$), which translates but does not rescale the cost matrix. The RD frontier is then traced by scaling $\rho$ over 50 log-spaced values of the inverse temperature $\lambda \in [10^{-3}, 10^{3}]$ (Section~\ref{sec:confusion_matrices}), computing the rate $R_k = I(X;Y)$ in bits and the expected distortion $D_k = \mathbb{E}[\rho(X,Y)]$ under an empirical class prior, which is uniform for the balanced GEN design. Local slopes $s_k$ are finite differences taken after sorting by $D$ and removing duplicate $D$ values, with $\beta \equiv \mathrm{median}_k\, s_k$ and $\kappa \equiv \mathrm{Var}_k(s_k)$.

The adequacy of this specification is bounded by the recoverability analysis in simulation (Appendix~\ref{sec:sim_secondary_diagnostics}), which compares the recovered symmetric and antisymmetric components of $\rho$ against known ground truth --- the antisymmetric component is recovered only under broad--weak structure --- and reports strict-recovery pass rates across generation regimes. Sensitivity to the $\lambda$ grid was assessed by re-deriving all three signatures over four alternative grids (ranges $[10^{-2},10^{2}]$ and $[10^{-4},10^{4}]$; 25 and 100 grid points) on a stratified subsample of 196 blocks. AUC is essentially invariant: its rank ordering across blocks is identical to the canonical grid in every case (Spearman $\rho = 1.000$) and group means shift by less than 4\%. $\beta$ and $\kappa$ are rank-stable but range-dependent in magnitude (Spearman $\rho = 0.72$--0.96 and $0.92$--0.98 respectively), with $\kappa$ varying by up to four orders of magnitude across grid ranges. This is expected rather than anomalous: $\beta$ is the median and $\kappa$ the variance of local slopes along the traced segment, so both are defined relative to the swept range, whereas AUC integrates the full curve. Comparisons involving $\beta$ and $\kappa$ magnitudes should therefore be read as conditional on the stated grid. All comparisons reported are made at the single canonical grid and are internally consistent. What does not transfer is the absolute scale of $\beta$ and $\kappa$, which should not be compared against values computed under a different grid specification.

Even though the prior rates themselves were not varied by re-fitting, the $\lambda$ sweep bounds the dominant part of this sensitivity. Blahut--Arimoto depends on $\rho$ only through $\exp(-\lambda\rho)$, so rescaling $\hat\rho$ is exactly equivalent to shifting the $\lambda$ range. Because the prior is common to every fit, any such rescaling applies identically to all blocks and leaves the rank- and slope-based between-block comparisons we report unchanged, consistent with the invariance observed above. What remains untested is the \emph{relative} weighting of the three prior components, which alters the shape of $\hat\rho$ rather than its scale.

\subsection{Inference and Collapsed-Channel Exclusion}
\label{app:inference}
\begin{figure*}[tbp]
  \centering
  \includegraphics[width=\textwidth]{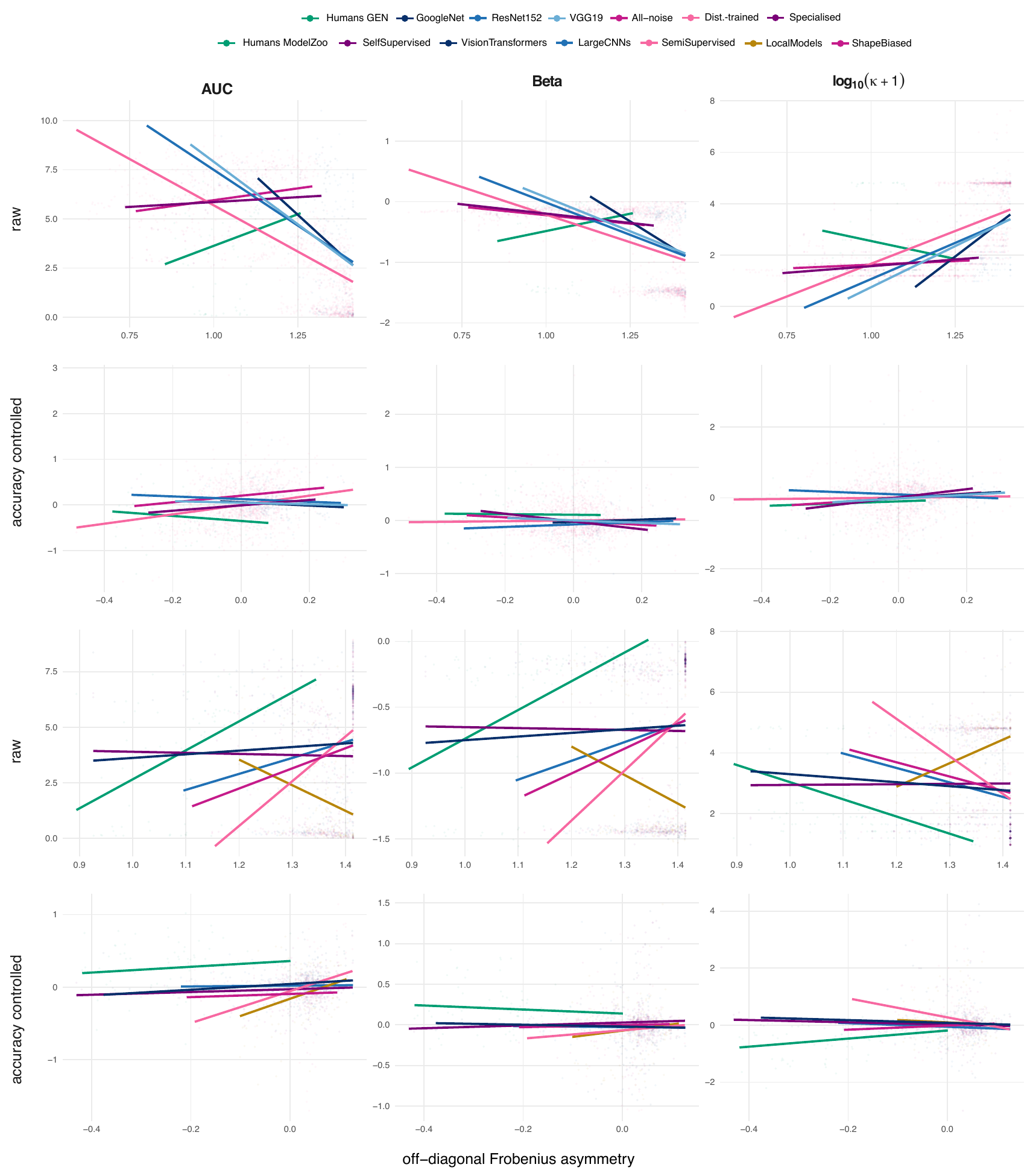}
  \caption{\textbf{Asymmetry--RD coupling before and after accuracy control, in both corpora.}
Each point is one block (experiment$\times$condition$\times$model); lines are per-group linear fits. The $x$-axis is off-diagonal Frobenius asymmetry $A_F^{\mathrm{off}}$.
\textbf{Rows 1--2:} GEN benchmark ($n=1610$ blocks across 115 experiment$\times$condition cells); the \emph{Raw} row reproduces Fig.~\ref{fig:asym_decomposition_2} with the \emph{Accuracy-controlled} row added.
\textbf{Rows 3--4:} ModelZoo architectures ($n=1135$ blocks across 73 cells).
In the \emph{Raw} rows both axes are untransformed. In the \emph{Accuracy-controlled} rows both axes are demeaned within (experiment, condition) and residualized on accuracy, so the fitted slope in each panel is the slope reported in Tables~\ref{tab:acc_controlled_gen} and~\ref{tab:acc_controlled_mz}. Panel scales are free, since the raw and residualized ranges differ.
Before accuracy control, greater asymmetry coincides with lower efficiency, steeper slope and higher curvature in most model families, but not in humans or in the robustness-trained regimes, whose couplings run the other way (Table~\ref{tab:asym_rd_by_group_A2}A). After accuracy control most couplings flatten, and those that remain are positive for AUC: the Distortion-trained regime in GEN and the local-receptive-field models in ModelZoo. The two corpora are fit separately, because demeaning within (experiment, condition) across a pooled corpus alters the GEN estimates.}
  \label{fig:asym_rd_before_after}
\end{figure*}

Within-block demeaning removes block-level means, but rows within a block remain non-independent, because a block contains many model instances evaluated on the same stimuli, and the Distortion-trained regime alone contributes roughly ten rows per block. Ordinary least squares treats these as independent observations and understates the standard errors for the groups with many instances per block, while the pooled residual variance -- dominated by the largest group -- overstates them for the smallest. We therefore report cluster-robust standard errors with (experiment, condition) blocks as clusters, using the bias-reduced CR2 estimator with Satterthwaite degrees of freedom \citep{pustejovsky2018small}, and draw the conclusions in the main text from these. OLS values are given alongside in Tables~\ref{tab:acc_controlled_gen} and~\ref{tab:acc_controlled_mz} for comparison. The two schemes agree on every qualitative conclusion except the Specialised $\beta$ association, which survives correction under CR2 ($p_{\mathrm{BH}}=0.037$) but not under OLS ($p_{\mathrm{BH}}=0.31$), but that association is in turn not robust to the $\lambda$ grid (Appendix~\ref{app:map_prior}), so we do not rely on it. Satterthwaite degrees of freedom range from 7.6 to 39.2, so inference is weakest for the groups with fewest independent blocks (ResNet-152, VGG-19) regardless of scheme. Figure~\ref{fig:asym_rd_before_after} shows the same relationships graphically for both corpora.

Collapsed channels are excluded as described in Appendix~\ref{app:collapse_filter}, computed from the row-normalized conditional distribution of each stored confusion matrix: 40 of 1650 GEN blocks are excluded (Distortion-trained 3.11\%, ResNet-152 and VGG-19 1.23\% each, all other groups 0\%), leaving $n=1610$. This exclusion matters most for $\beta$, which is the signature most sensitive to near-degenerate channels. The excluded blocks occur at near-chance distortion levels, where a system's responses collapse onto a single class. In 24 of the 40 this produces a numerically non-monotone frontier segment and a positive $\beta$ (Appendix~\ref{sec:rd_nonmonotone}). Retaining them changes the Distortion-trained $\beta$ slope from $0.07$ to $0.46$

The two corpora are fit separately. Demeaning operates within (experiment, condition), so fitting GEN and ModelZoo jointly pools their rows into shared blocks and alters the GEN estimates. The GEN analysis reported here is the one that uses GEN blocks alone. The distinction is consequential. Fitted separately, local models (BagNet-9/33) retain a significant accuracy-controlled association between asymmetry and RD efficiency (slope $+2.53$, $p_{\mathrm{BH}}=0.002$ under CR2; Table~\ref{tab:acc_controlled_mz}), which the joint fit does not recover. No other modern family retains an association on any signature.

\begin{table}[t]
\caption{\textbf{Accuracy-controlled asymmetry--RD slopes (GEN).} Per-group within-block slopes of each RD signature on off-diagonal Frobenius asymmetry, controlling for accuracy within (experiment, condition) blocks ($n=1610$ blocks after collapsed-channel exclusion; 115 demeaning blocks). Slopes are identical under both inference schemes; only the standard errors differ. OLS treats blocks' rows as independent; CR2 is the bias-reduced cluster-robust estimator with blocks as clusters and Satterthwaite degrees of freedom. $p_{\mathrm{BH}}$ is corrected across the seven groups within each signature. Conclusions in the main text follow the CR2 column. Of the two associations significant under CR2, only the Distortion-trained AUC coupling is also robust to the $\lambda$ grid; the Specialised $\beta$ association is not (Appendix~\ref{app:map_prior}).}
\label{tab:acc_controlled_gen}
\centering\small
\setlength{\tabcolsep}{5pt}
\begin{tabular}{lrrrrrr}
\toprule
& & \multicolumn{2}{c}{OLS} & \multicolumn{3}{c}{Cluster-robust (CR2)} \\
\cmidrule(lr){3-4}\cmidrule(lr){5-7}
Group & Slope & SE & $p_{\mathrm{BH}}$ & SE & df & $p_{\mathrm{BH}}$ \\
\midrule
\multicolumn{7}{l}{\textit{Efficiency (AUC)}}\\
Humans             & $-0.886$ & 0.479 & 0.225 & 0.478 & 22.4 & 0.147 \\
GoogLeNet          & $0.722$  & 0.770 & 0.418 & 0.558 & 20.2 & 0.292 \\
ResNet-152         & $-0.458$ & 0.498 & 0.418 & 0.368 & 7.6  & 0.292 \\
VGG-19             & $0.087$  & 0.642 & 0.892 & 0.532 & 8.5  & 0.874 \\
Distortion-trained & $\mathbf{1.024}$ & 0.117 & $4.2{\times}10^{-17}$ & 0.297 & 37.9 & $\mathbf{0.010}$ \\
Specialised        & $0.585$  & 0.501 & 0.418 & 0.324 & 22.9 & 0.147 \\
All-noise          & $0.694$  & 0.468 & 0.323 & 0.324 & 16.1 & 0.147 \\
\addlinespace
\multicolumn{7}{l}{\textit{Slope ($\beta$)}}\\
Humans             & $-0.495$ & 0.371 & 0.640 & 0.239 & 22.4 & 0.154 \\
GoogLeNet          & $0.141$  & 0.598 & 0.814 & 0.465 & 20.2 & 0.765 \\
ResNet-152         & $0.123$  & 0.386 & 0.814 & 0.133 & 7.6  & 0.535 \\
VGG-19             & $-0.334$ & 0.498 & 0.704 & 0.244 & 8.5  & 0.360 \\
Distortion-trained & $0.067$  & 0.091 & 0.704 & 0.188 & 37.9 & 0.765 \\
Specialised        & $-0.785$ & 0.389 & 0.306 & 0.255 & 22.9 & 0.037 \\
All-noise          & $-0.347$ & 0.363 & 0.704 & 0.176 & 16.1 & 0.154 \\
\addlinespace
\multicolumn{7}{l}{\textit{Curvature ($\log_{10}(\kappa{+}1)$)}}\\
Humans             & $0.833$  & 0.614 & 0.614 & 0.382 & 22.4 & 0.132 \\
GoogLeNet          & $0.588$  & 0.989 & 0.674 & 0.674 & 20.2 & 0.496 \\
ResNet-152         & $-0.278$ & 0.639 & 0.674 & 0.241 & 7.6  & 0.496 \\
VGG-19             & $0.346$  & 0.824 & 0.464 & 0.413 & 8.5  & 0.496 \\
Distortion-trained & $0.107$  & 0.150 & 0.674 & 0.287 & 37.9 & 0.712 \\
Specialised        & $1.322$  & 0.643 & 0.279 & 0.497 & 22.9 & 0.098 \\
All-noise          & $0.627$  & 0.600 & 0.674 & 0.305 & 16.1 & 0.132 \\
\bottomrule
\end{tabular}
\end{table}

\begin{table}[t]
\caption{\textbf{Accuracy-controlled asymmetry--RD slopes (ModelZoo).} As Table~\ref{tab:acc_controlled_gen}, fit separately on the ModelZoo corpus ($n=1135$ blocks, 73 demeaning blocks) with ModelZoo humans as reference. The corpora are fit separately because within-block demeaning across a pooled corpus alters the GEN estimates.}
\label{tab:acc_controlled_mz}
\centering\small
\setlength{\tabcolsep}{5pt}
\begin{tabular}{lrrrrrr}
\toprule
& & \multicolumn{2}{c}{OLS} & \multicolumn{3}{c}{Cluster-robust (CR2)} \\
\cmidrule(lr){3-4}\cmidrule(lr){5-7}
Group & Slope & SE & $p_{\mathrm{BH}}$ & SE & df & $p_{\mathrm{BH}}$ \\
\midrule
\multicolumn{7}{l}{\textit{Efficiency (AUC)}}\\
Humans (ModelZoo)   & $0.370$ & 0.331 & 0.461 & 0.518 & 27.4 & 0.673 \\
Vision Transformers & $0.400$ & 0.247 & 0.245 & 0.298 & 12.2 & 0.409 \\
Large CNNs (BiT-M)  & $0.057$ & 0.333 & 0.865 & 0.255 & 27.6 & 0.825 \\
Semi-supervised     & $2.075$ & 0.457 & $2.2{\times}10^{-5}$ & 0.981 & 14.2 & 0.184 \\
Self-supervised     & $0.201$ & 0.228 & 0.528 & 0.161 & 13.4 & 0.409 \\
Shape-biased (SIN)  & $0.114$ & 0.337 & 0.858 & 0.272 & 28.7 & 0.792 \\
Local models        & $\mathbf{2.525}$ & 0.414 & $1.0{\times}10^{-8}$ & 0.635 & 39.2 & $\mathbf{0.002}$ \\
\addlinespace
\multicolumn{7}{l}{\textit{Slope ($\beta$)}}\\
Humans (ModelZoo)   & $-0.286$ & 0.378 & 0.937 & 0.369 & 27.4 & 0.780 \\
Vision Transformers & $-0.083$ & 0.282 & 0.937 & 0.185 & 12.2 & 0.903 \\
Large CNNs (BiT-M)  & $-0.049$ & 0.381 & 0.937 & 0.317 & 27.6 & 0.903 \\
Semi-supervised     & $0.456$  & 0.523 & 0.937 & 0.580 & 14.2 & 0.780 \\
Self-supervised     & $0.154$  & 0.260 & 0.937 & 0.126 & 13.4 & 0.780 \\
Shape-biased (SIN)  & $0.031$  & 0.386 & 0.937 & 0.248 & 28.7 & 0.903 \\
Local models        & $0.627$  & 0.473 & 0.937 & 0.554 & 39.2 & 0.780 \\
\addlinespace
\multicolumn{7}{l}{\textit{Curvature ($\log_{10}(\kappa{+}1)$)}}\\
Humans (ModelZoo)   & $1.518$  & 0.743 & 0.145 & 0.684 & 27.4 & 0.206 \\
Vision Transformers & $-0.521$ & 0.554 & 0.394 & 0.406 & 12.2 & 0.307 \\
Large CNNs (BiT-M)  & $-0.637$ & 0.748 & 0.394 & 0.502 & 27.6 & 0.307 \\
Semi-supervised     & $-3.177$ & 1.028 & 0.014 & 1.591 & 14.2 & 0.206 \\
Self-supervised     & $-0.505$ & 0.511 & 0.394 & 0.275 & 13.4 & 0.206 \\
Shape-biased (SIN)  & $0.660$  & 0.758 & 0.394 & 0.595 & 28.7 & 0.307 \\
Local models        & $-0.827$ & 0.930 & 0.394 & 0.799 & 39.2 & 0.307 \\
\bottomrule
\end{tabular}
\end{table}

\subsection{Asymmetry Metric Implementation}
\label{app:asym_impl}
Breadth, strength, and the Frobenius index are computed on the row-normalized confusion matrix $C$ (Section~\ref{sec:asym_metrics}). A class pair $i<j$ is counted as asymmetric when $|C_{ij}-C_{ji}|>\varepsilon$, with $\varepsilon=10^{-12}$ for all empirical analyses and $\varepsilon=10^{-6}$ for the simulation summaries (Appendix~\ref{sec:sim_results_appx}). Because $f_{\mathrm{pairs}}$ counts pairs exceeding $\varepsilon$, it increases with the number of trials per cell, and the subsampling control in Appendix~\ref{app:per_observer} bounds that contribution. $A_F^{\mathrm{off}}$ carries no threshold and is the asymmetry predictor in every RD analysis reported here, so none of the asymmetry--RD results inherits this sensitivity.

\begin{figure*}[tbp]
  \centering
  \includegraphics[width=\textwidth]{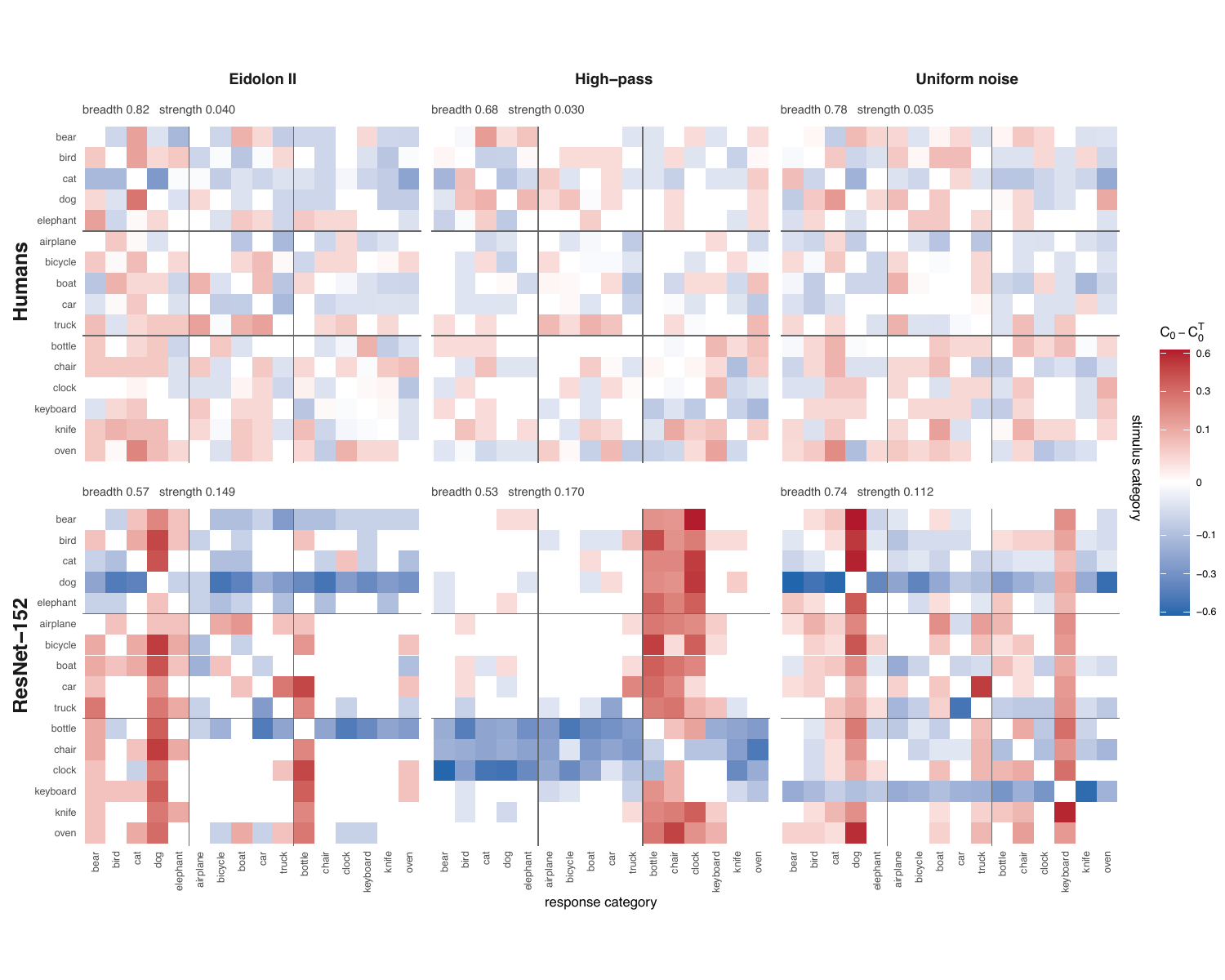}
  \caption{\textbf{Representative directional confusion maps.} Off-diagonal asymmetry maps $C_0 - C_0^{\mathsf{T}}$ for humans (top) and ResNet-152 (bottom) under three perturbation families, where $C_0$ is the row-normalized confusion matrix with the diagonal removed. Cell $(i,j)$ is positive (red) when category $i$ is misread as $j$ more often than the reverse, and negative (blue) when $i$ receives more than it sends, so a predominantly blue row marks a class acting as an error sink. Classes are ordered by superordinate category (animals, vehicles, objects; grey lines). Breadth and strength are printed above each panel, computed as in Appendix~\ref{app:asym_impl}. The color scale is signed-square-root so that the weak human structure stays visible alongside the strong ANN structure. Humans show pale asymmetries distributed across nearly all pairs. ResNet-152 resolves into a few strong receiving columns whose identity changes across conditions --- \emph{dog} under Eidolon~II, \emph{bottle}/\emph{chair}/\emph{clock} under high-pass, \emph{dog} and \emph{keyboard} under uniform noise --- the per-matrix form of the modal-receiver instability reported in Section~\ref{sec:results_robust}. Because the two rows are matched on stimuli and conditions, the human panels serve directly as the diffuse reference for the concentrated ANN panels.}
  \label{fig:confusion_asymmetry_K16}
\end{figure*}

\subsection{Per-Observer Analysis}
\label{app:per_observer}
To test whether the human profile is an artifact of aggregating observers, we analyzed per-condition individual-observer confusion matrices from the 5--6 observers with adequate coverage in the GEN human data, computed over the 4--6 perturbation families each observer completed. The strength half of the dissociation is robustly individual: every observer's mean strength (0.024--0.043) falls entirely below the baseline CNN and distortion-trained range in the same corpus (0.121--0.154), and every observer's Frobenius asymmetry sits at the low, human end. The specialised and all-noise regimes are the exceptions (0.030 and 0.025), and they are also the two groups statistically indistinguishable from humans on global asymmetry (Table~\ref{tab:group_tests_A7}). Individual breadth is lower than the pooled human value. A subsampling control indicates this is a trial-count effect, as thinning the pooled human matrix to a single observer's trial count reduces its breadth to the individual level (median 0.33 vs.\ 0.32 at matched per-condition density). Aggregate breadth should therefore be read as a population-level property, with the per-observer signature carried mainly by low strength. We find no evidence of distinct strategy subgroups, since cross-observer correlations of per-experiment breadth are uniformly high (0.90--0.99), and inter-observer breadth SD within an experiment (${\approx}0.07$) is small relative to the GEN human--model breadth gap ($0.712$ vs.\ $0.535$; Table~\ref{tab:asymmetry_appendix_summary_A1}). 

Sparsity does not separate the two populations at per-condition GEN density. Individual observers occupy 0.46--0.56 in fraction of empty off-diagonal cells, against $0.438$ for the pooled human matrix and 0.54--0.71 for the GEN ANN groups. We therefore do not interpret individual-level sparsity as human-like, and only strength and Frobenius asymmetry are cleanly individual.

\subsection{Error-Mass Concentration and Receiver Stability}
\label{app:sink}
If ANN sparsity reflected a single sink category, the number of classes receiving above-chance error mass would be close to one and the identity of the dominant receiver would be constant across perturbation conditions. Neither holds. Averaged over models, error mass above chance falls on 4.95 classes (humans: 6.07), and each model has 7--18 distinct top receivers across conditions, with modal-receiver consistency of 0.21--0.46 (humans: 0.73). Concentration is nonetheless higher than in humans: the single largest receiving class absorbs an excess of 0.281 over a density-matched null (humans: 0.086), and the top three classes absorb 0.63 of all error mass (humans: 0.41). Ranked by regime, top-1 concentration is 0.145 (all-noise), 0.148 (specialised), 0.178 (humans), 0.378 (Distortion-trained) and 0.412--0.446 (baseline CNNs).

\subsection{Label-Grouping Robustness}
\label{app:label_grouping}
We recomputed all asymmetry metrics under two relabelings of the same trials. The semantic grouping collapses the 16 GEN categories into 8 superordinate categories, but the direction and significance of all three effects are preserved (breadth: humans $0.92$ vs.\ ANNs $0.79$, $p=6\times10^{-13}$; strength: $0.048$ vs.\ $0.16$, $p=7\times10^{-5}$; Frobenius: $0.89$ vs.\ $1.15$, $p=5\times10^{-25}$). Because any single semantic grouping is one particular choice of partition, we additionally drew 50 random partitions of the 16 classes into 8 groups and recomputed the metrics for each; humans showed higher breadth and lower strength than ANNs in 50 of 50. Both analyses re-partition the same images, trials, responses and models, so they test robustness to how labels are grouped and not to the size of the label space, which is addressed in Appendix~\ref{app:imagenet1k}. Breadth and strength both increase under coarsening relative to their $K=16$ values (Table~\ref{tab:asymmetry_appendix_summary_A1}), so what is preserved is the direction of the human--ANN difference rather than the absolute magnitudes.

\section{Directional-Structure Ablations}
\label{app:ablation}
\paragraph{Design.}
To quantify how much the RD signatures depend on directional structure, we apply two ablations to a matched subset of GEN matrices --- humans, a baseline CNN (ResNet-152), and a specialised training-regime model, across four perturbation experiments (high-pass, low-pass, phase scrambling, rotation), 27 matrices each. Each ablated matrice was passed through the same MAP and frontier pipeline used for the originals. \emph{Symmetrization} mirrors one triangle of $C$ onto the other, the operation implicit whenever behavioral data are forced into a symmetric similarity space. Because the choice of triangle (upper vs. lower) is arbitrary, we apply both directions (mirror-lower and mirror-upper) and treat their disagreement as a diagnostic. The \emph{within-row off-diagonal shuffle} permutes the off-diagonal entries within each row, scrambling which classes receive a system's errors while leaving the diagonal --- and therefore accuracy and each class's total error mass --- unchanged. This is applied with three independent seeds per matrix. The five ablated variants per original matrix --- two mirror directions and three shuffle seeds --- give $81\times5=405$ ablated matrices in total. Comparisons are paired within matrix.

Both manipulations act on integer counts, so neither is as clean as its name suggests. Mirroring symmetrizes the counts but leaves row totals unequal, so the row-normalized channel retains part of its asymmetry: $49\%$ of $A_F^{\mathrm{off}}$ is removed for ResNet-152, against $73\%$ for humans and $82\%$ for the specialised model. The shuffle is close to magnitude-preserving for concentrated systems, where little symmetric structure remains for a permutation to destroy (ResNet-152: $A_F^{\mathrm{off}}$ $+1.9\%$ on average), but raises magnitude for broader ones (humans $+10.5\%$, specialised $+17.6\%$).

\paragraph{Signature normalization.}
The ablation signatures are computed on min--max normalized frontiers, with $R$ and $D$ each rescaled to $[0,1]$ within each curve, rather than on the raw frontiers used elsewhere in the paper. Symmetrization alters both the scale of the inferred frontier and its shape. Normalizing within curve removes the scale change, so the reported $\Delta\log_{10}(\kappa{+}1)$ isolates the change in trade-off shape, which is the quantity the ablation is about. Signature values in this section are therefore not comparable to those in Table~\ref{tab:acc_controlled_gen} or Fig.~\ref{fig:asym_decomposition_2}. Computed on raw frontiers the same comparisons are not significant ($p=0.16$--$0.50$). As expected, at raw scale, $\kappa$ spans nine orders of magnitude, so a change in frontier shape produces a negligible change in $\log_{10}\kappa$.

\paragraph{Control: the effect is not driven by asymmetry magnitude.}
Across all 405 ablated matrices, the change in $A_F^{\mathrm{off}}$ accounts for under $2\%$ of the variance in $\Delta\log_{10}(\kappa{+}1)$ (adjusted $R^2 = 0.015$), and the ordering runs against a magnitude account: the shuffle changes ResNet-152's magnitude least and its curvature most, while symmetrization removes the smallest share of its asymmetry and produces its largest curvature shift. Because the 27 matrices per system span multiple distortion levels within four experiments, they are not independent samples. The reported Wilcoxon statistics are paired comparisons across matched matrices and should be read as descriptive effect sizes rather than tests over independent observations.

\paragraph{Results.}
Symmetrization raises curvature in close to every matrix (fraction with positive $\Delta\log_{10}(\kappa{+}1)$: $0.91$, $0.89$, $0.93$), and the size of the rise orders the three systems as their error concentration does: specialised $0.16$, humans $0.24$, ResNet-152 $0.59$ (medians; all pairwise Wilcoxon $p=1.7\times10^{-6}$ to $0.029$). The disagreement between the two arbitrary mirror directions --- the directional information a symmetry assumption discards --- follows the same ordering ($0.07$, $0.16$, $0.22$). However, the shuffle separates the systems more sharply still. It changes curvature only slightly for humans and the specialised model ($-0.06$ in both) while shifting ResNet-152 by $+0.35$ ($p=1.1\times10^{-5}$). Efficiency moves in the opposite pattern --- symmetrization costs humans the most AUC ($-6.9$ to $-10.3\%$) and ResNet-152 the least ($-1.1$ to $-6.4\%$). Full values are in Table~\ref{tab:ablation_variants}.

\begin{figure}[tbp]
  \centering
  \includegraphics[width=0.85\linewidth]{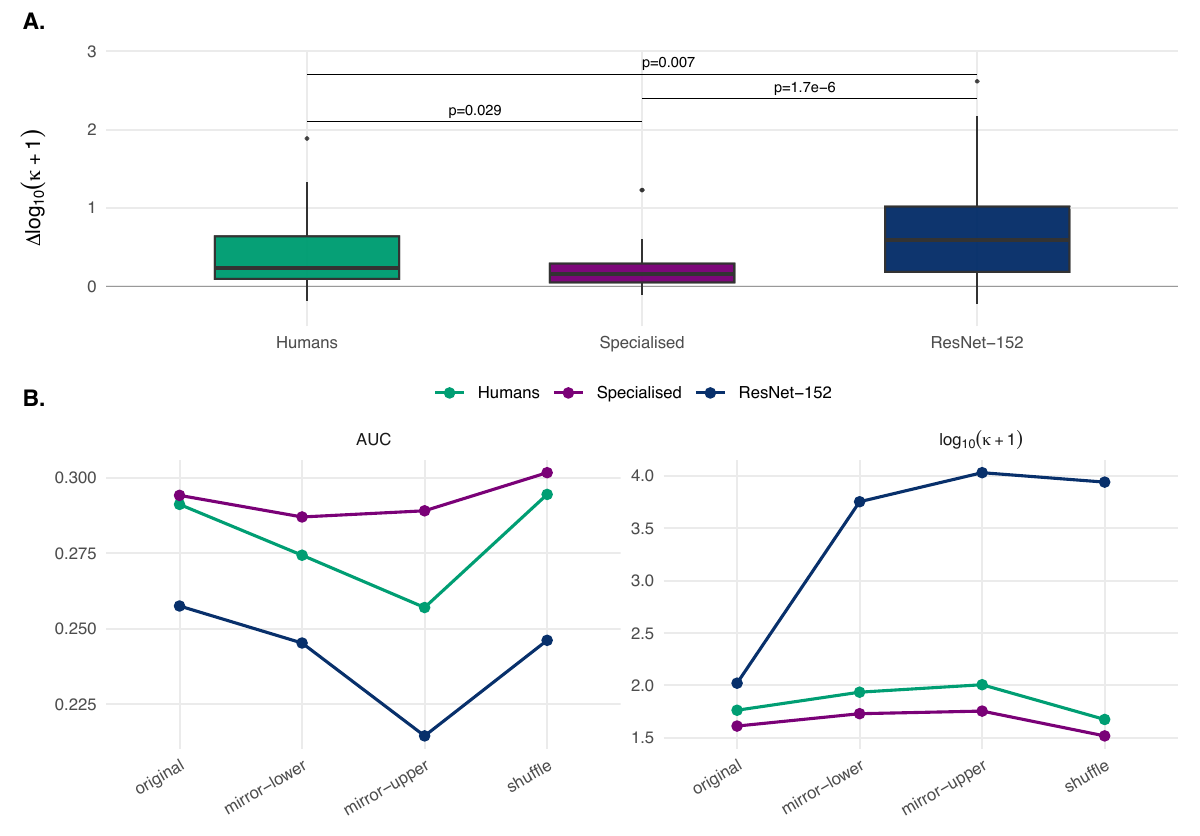}
  \caption{\textbf{Removing directional structure reshapes RD geometry most for the system with the most concentrated error mass.} Humans, a specialised-regime model and a baseline CNN (ResNet-152) under the original matrices and three ablations: mirroring the lower triangle onto the upper (\emph{mirror-lower}), the reverse (\emph{mirror-upper}), and a within-row off-diagonal shuffle that scrambles the directional target of errors while holding the diagonal --- and hence accuracy and per-class error magnitude --- fixed (\emph{shuffle}). \textbf{(A)}~Curvature fragility: the per-matrix increase in $\log_{10}(\kappa{+}1)$ under symmetrization, with two-sided Wilcoxon comparisons. The ordering is specialised $<$ humans $<$ ResNet-152. \textbf{(B)}~Median signature value per system at each manipulation, for efficiency (AUC, left) and curvature ($\log_{10}\kappa$, right). The two panels report different statistics: (A) is the median of per-matrix \emph{differences}, whereas (B) plots the median signature at each condition, so the vertical rise visible in (B) is a difference of medians and exceeds the median difference in (A) under the skewed $\kappa$ distribution. All points aggregate four perturbation experiments (high-pass, low-pass, phase scrambling, rotation; 27 matrices each). The baseline CNN separates from humans and the specialised model under every ablation, and the two arbitrary mirror directions disagree with each other most for it: symmetrization is least well-defined for the systems carrying the most directional structure.}
  \label{fig:ablation_signatures}
\end{figure}

\begin{table}[t]
\caption{\textbf{Directional-structure ablations.} Effects of removing directional information from GEN confusion matrices, for humans, a specialised single-distortion model, and a baseline CNN, across four perturbation experiments (27 matrices per system). Rows are ordered by fragility. \textbf{(A)} $\log_{10}\kappa$ by variant, median [IQR] over matrices. \textbf{(B)} Ablation effects. Inflation is the per-matrix increase in $\log_{10}\kappa$ under symmetrization, pooled over both mirror directions ($n=54$); ``frac.\ pos.'' is the fraction of matrices in which it increases; ``mirror disagr.'' is the median absolute difference in $\log_{10}\kappa$ between the two arbitrary symmetrizations. Shuffle columns give the paired change relative to the original matrix, with Wilcoxon $p$-values. Changes in $\beta$ are consistent in direction (steeper under symmetrization in every system) but are not reported numerically, because $\beta$ is a median of local slopes that can approach zero, making relative changes unstable.}
\label{tab:ablation_variants}
\centering\small
\setlength{\tabcolsep}{5pt}

\textbf{(A) $\log_{10}\kappa$ by variant}\par\vspace{0.5ex}
\begin{tabular}{lcccc}
\toprule
System & Original & Mirror-lower & Mirror-upper & Shuffle \\
\midrule
Specialised  & 1.61 [1.50--1.83] & 1.73 [1.69--1.94] & 1.75 [1.69--1.95] & 1.52 [1.46--1.72] \\
Humans       & 1.76 [1.60--2.26] & 1.94 [1.73--2.84] & 2.01 [1.73--3.24] & 1.67 [1.51--2.21] \\
ResNet-152   & 2.02 [1.77--3.94] & 3.75 [1.87--4.66] & 4.03 [2.07--4.54] & 3.94 [1.80--4.66] \\
\bottomrule
\end{tabular}

\vspace{1.2ex}

\textbf{(B) Ablation effects}\par\vspace{0.5ex}
\begin{tabular}{lccccccc}
\toprule
& \multicolumn{3}{c}{Symmetrization} & & \multicolumn{2}{c}{Shuffle} \\
\cmidrule(lr){2-4}\cmidrule(lr){6-7}
System & \shortstack{$\kappa$ inflation\\median (IQR)} & \shortstack{frac.\\pos.} & \shortstack{mirror\\disagr.} & \shortstack{AUC\\change} & \shortstack{$\Delta\log_{10}\kappa$\\($p$)} & \shortstack{AUC\\change ($p$)} \\
\midrule
Specialised & 0.16 (0.24) & 0.91 & 0.07 & $-2.8$ to $-4.6\%$  & $-0.06$ ($0.0055$) & $+0.5\%$ ($0.20$) \\
Humans      & 0.24 (0.54) & 0.89 & 0.16 & $-6.9$ to $-10.3\%$ & $-0.06$ ($0.90$)   & $+3.1\%$ ($0.017$) \\
ResNet-152  & \textbf{0.59} (0.83) & 0.93 & \textbf{0.22} & $-1.1$ to $-6.4\%$ & $\mathbf{+0.35}$ ($1.1{\times}10^{-5}$) & $+3.0\%$ ($0.059$) \\
\bottomrule
\end{tabular}
\end{table}

\section{Extension to Modern Architectures}
\label{sec:architectures_appx}
\subsection{Model selection and data.}
The main analyses use matched human psychophysics and ANN behavior from the GEN benchmark repository \citep{geirhos2018generalisation}, which provides $\sim$83k human trials across 12 perturbation families at graded distortion levels for $K = 16$ ImageNet-derived categories. Architecture choice in the main analysis was constrained by the requirement for matched human psychophysics data at sufficient trial density. Only models for which per-cell confusion counts are dense enough to reliably estimate directional asymmetry at the class-pair level are included. This limits the main analysis to three baseline CNNs (GoogLeNet, ResNet-152, VGG-19) and three robustness training regimes (Distortion-trained, Specialised, All-noise).

To assess whether the global asymmetry dissociation extends to modern architectures not included in the GEN repository, we additionally analyzed a broad selection of 15 architectures spanning six architecture families:
\begin{itemize}
    \item     \textbf{Large-scale supervised models}: BiT models (bitm-resnetv2-152x2, bitm-resnetv2-152x4 \citep{kolesnikov2020big})
    \item \textbf{Transformers and vision--language models}: Vision Transformer ViT Large Patch16-224 \citep{dosovitskiy2021imageworth16x16words}, SWAG-ViT-L/16-IN1K \citep{singh2022swag}, CLIP \cite{radford2021clip}
    \item \textbf{Architectural inductive-bias probes (LocalModels)}: BagNet-9 and BagNet-33 \citep{brendel2019approximating}
    \item \textbf{Robustness/shape-biased models}: standard ResNet-50 CNNs trained on Stylized ImageNet (ResNet50 SIN) and a mixture of Stylized and regular ImageNet ResNet50 (SIN+IN) \citep{geirhos2019imagenet}
    \item \textbf{Data rich semi-supervised models}: semi-weakly supervised pretrained ResNeXt-101 32x16d SWSL \citep{mahajan2018exploring}, Noisy Student pretraining (efficientnet-l2-noisy-student-475 \cite{xie2020noisystudent})
    \item \textbf{Self-supervised / contrastive models}: Simple Framework for Contrastive Learning models (SimCLR resnet50x1, SimCLR resnet50x4 \cite{chen2020simple}), MoCo v2 \cite{chen2020mocov2}, InfoMin \cite{tian2020infomin}
\end{itemize}
Model outputs were obtained from the ModelZoo repository \citep{geirhos2021partial} using the same stimulus set, enabling direct comparison of asymmetry structure under identical perturbation conditions, with a second set of human data ($\sim$85k psychophysics trials).

\begin{figure*}[tbp]
  \centering
  \includegraphics[width=\textwidth]{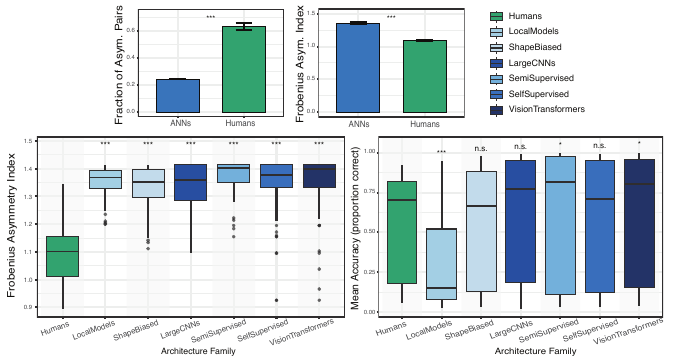}
  \caption{\textbf{Modern architectures are more asymmetric than humans at every accuracy level.}
\textbf{Top row:} Fraction of asymmetric class pairs (left) and Frobenius asymmetry index (right) comparing humans against all 15 modern model architectures. Error bars show s.e.m. and significance marks correspond to two-sided Wilcoxon rank-sum tests on block-wise summaries.
\textbf{Bottom row:} Frobenius asymmetry index (left) and mean classification accuracy (right) by architecture family. Each point summarizes one block (unique experiment$\times$condition$\times$model). Boxes show the distribution across blocks. Significance marks are based on planned Wilcoxon comparisons of each family against humans with BH--FDR correction.}
\label{fig:asym_modelzoo}
\end{figure*}

\subsection{Analytical approach and sparsity caveat}
Asymmetry metrics are computed identically to the main analysis 
(Section~\ref{sec:asym_metrics}), using block-wise summaries (one value per unique experiment$\times$condition$\times$model) on row-normalized confusion matrices. Group comparisons use two-sided Wilcoxon rank-sum tests with BH--FDR correction across planned comparisons against humans.

An important limitation of the ModelZoo repository data is that confusion matrices are sparser than in the GEN repository, since these models were not evaluated with matched per-stimulus trial counts, resulting in fewer observations per confusion matrix cell. The GEN repository confusion matrices contain on average 85 non-zero asymmetric class pairs out of 120 possible for humans ($K = 16$; 71\%) and 64 for ANNs (54\%), providing sufficient density for the breadth--strength decomposition. By contrast, the ModelZoo ANN confusion matrices contain on average only 29 pairs (24\%), less than half the GEN ANN coverage despite identical $K = 16$ categories and stimulus conditions. ModelZoo human data achieves 
comparable coverage to the GEN benchmark (76 pairs; 63\%) confirming that the drop in pair coverage is specific to the ModelZoo ANN confusion matrices. This sparsity simultaneously deflates estimates of asymmetry breadth ($f_\text{pairs}$), because many off-diagonal 
cells are exactly zero, and inflates conditional magnitude ($\bar{\Delta}$), because only the largest asymmetries survive the sparsity filter. These two metrics should therefore be interpreted with caution for the per-condition ModelZoo matrices. The Frobenius asymmetry index, which aggregates over all off-diagonal entries and is substantially less sensitive to the proportion of zero cells, is robust to this issue and is therefore the primary metric reported in the per-condition analysis below. Pooling each perturbation family across its severity levels raises density enough for the full breadth--strength decomposition (see Appendix~\ref{sec:architectures_pooled} for this analysis).

\subsection{Results}
All six ModelZoo architecture families (15 architectures) remain significantly more asymmetric than humans on the Frobenius index (all $p < 0.001$, BH--FDR corrected; \textbf{Appendix Fig.~\ref{fig:asym_modelzoo}}, bottom-left), replicating the global asymmetry dissociation observed in the main analysis across a substantially broader range of architectures. This result holds regardless of training paradigm, including self-supervised (SimCLR, MoCoV2, InfoMin), semi-supervised (ResNeXt101-SWSL, EffNet-L2), shape-biased (ResNet50-SIN), and large-scale supervised models (BiT-M). Critically, it also holds for Vision Transformers and CLIP, which represent a qualitatively different architectural class from the CNNs evaluated in the main analysis. CLIP and SWAG-ViT-L/16 are trained with language or hashtag supervision rather than ImageNet labels alone, yet both fall within the transformer family's asymmetry profile --- one of the two least broad families --- giving no indication that multimodal supervision by itself reshapes directional organization.

The accuracy profile across architecture families is heterogeneous 
(\textbf{Appendix Fig.~\ref{fig:asym_modelzoo}}, bottom-right). Local Models (BagNet-9, BagNet-33) are significantly less accurate than humans, while Vision Transformers and semi-supervised models are significantly more accurate. The remaining families do not differ from humans. Yet all families are significantly more asymmetric than humans, confirming a double dissociation between asymmetry structure and accuracy that extends beyond the main analysis to modern architectures.

The main analyses in Sections~\ref{sec:results_asym} and~\ref{sec:results_rd} use the GEN data because it provides directly matched human psychophysics under identical stimulus conditions at per-condition density. The ModelZoo systems are analyzed separately in Appendix~\ref{sec:architectures_pooled}, where pooling across severity levels supports the breadth--strength decomposition. The accuracy-controlled RD analysis on the extended model set is per-condition and appears in Appendix~\ref{app:inference} (Table~\ref{tab:acc_controlled_mz}, Fig.~\ref{fig:asym_rd_before_after}).

\begin{table}[t]
\caption{\textbf{Modern architectures on pooled ModelZoo matrices.} Mean breadth ($f_{\mathrm{pairs}}$), strength ($\bar{\Delta}$), off-diagonal Frobenius asymmetry ($A_F^{\mathrm{off}}$), fraction of empty off-diagonal cells, and accuracy by architecture family, with humans as the reference row. The metrics are defined in Appendix~\ref{app:asym_impl}. Matrices are pooled per perturbation family across severity levels, giving $n=12$ blocks per family, and collapsed channels are excluded. Values are therefore not comparable to the per-condition analyses in Table~\ref{tab:asymmetry_appendix_summary_A1} and Fig.~\ref{fig:asym_modelzoo}. Pooled accuracies in particular are trial-weighted and run roughly $0.07$ above the per-condition means in that figure. Block-level distributions and significance against humans are in Fig.~\ref{fig:modelzoo_breadth_strength}.}
\label{tab:modelzoo_families}
\centering\small
\begin{tabular}{lccccc}
\toprule
Family & Breadth & Strength & Frobenius & Off-diag.\ zeros & Acc. \\
\midrule
Humans & 0.923 & 0.022 & 0.882 & 0.140 & 0.62 \\
Vision Transformers & 0.469 & 0.059 & 1.325 & 0.691 & 0.71 \\
Semi-supervised & 0.451 & 0.060 & 1.333 & 0.716 & 0.70 \\
Self-supervised & 0.548 & 0.064 & 1.297 & 0.634 & 0.65 \\
Large CNNs (BiT-M) & 0.578 & 0.050 & 1.238 & 0.588 & 0.67 \\
Shape-biased (SIN) & 0.648 & 0.055 & 1.223 & 0.531 & 0.62 \\
Local models (BagNet) & 0.720 & 0.090 & 1.283 & 0.447 & 0.39 \\
\bottomrule
\end{tabular}
\end{table}

\subsection{Breadth--strength decomposition on pooled matrices}
\label{sec:architectures_pooled}
\begin{figure}[tbp]
  \centering
  \includegraphics[width=\linewidth]{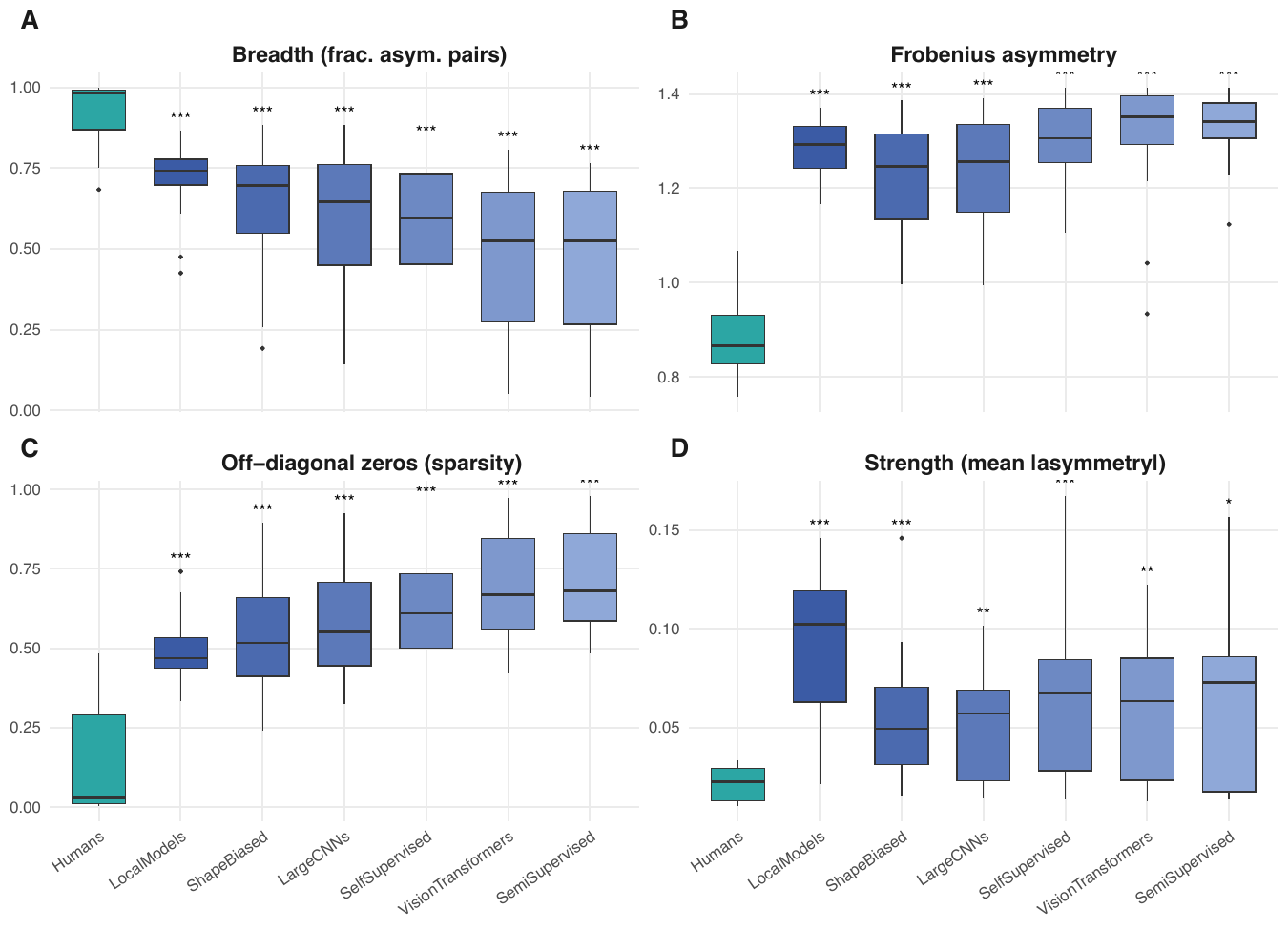}
  \caption{\textbf{Humans and modern architectures dissociate on all four asymmetry measures.} Block-level distributions for humans and six ModelZoo architecture families: \textbf{(A)}~breadth (fraction of asymmetric class pairs), \textbf{(B)}~Frobenius asymmetry $A_F^{\mathrm{off}}$, \textbf{(C)}~off-diagonal sparsity (fraction of empty off-diagonal cells), and \textbf{(D)}~strength (mean $|C_{ij}-C_{ji}|$ among asymmetric pairs). Each box summarizes $n=12$ blocks, one per shared parametric experiment. Matrices are pooled per perturbation family across severity levels, so values are not comparable to the per-condition analyses in Fig.~\ref{fig:asym_modelzoo} and Table~\ref{tab:asymmetry_appendix_summary_A1}. Every family is less broad, sparser, more Frobenius-asymmetric and higher in strength than humans. These are the 24 family$\times$metric comparisons reported in Section~\ref{sec:results_robust}. Stars denote planned two-sided Wilcoxon rank-sum comparisons against humans with BH--FDR correction ($^{*}p<0.05$, $^{**}p<0.01$, $^{***}p<0.001$). Family means are given in Table~\ref{tab:modelzoo_families}.}
  \label{fig:modelzoo_breadth_strength}
\end{figure}

To obtain the density required for the breadth--strength decomposition, we pooled each perturbation family across its severity levels, restricted to the 12 parametric experiments that humans and ModelZoo models share, yielding ${\sim}77$ trials per class for models and ${\sim}308$ for humans. Metrics are computed as in Section~\ref{sec:asym_metrics}. Per-family Wilcoxon tests compare $n=12$ family blocks against $n=12$ human blocks, but the pooled comparison against the benchmark-human-like models ($n=84$ vs.\ $12$) is better powered and is the one emphasized in the main text. Family means are given in Table~\ref{tab:modelzoo_families} and block-level distributions in Fig.~\ref{fig:modelzoo_breadth_strength}.

\section{Error Concentration at $K=1000$}
\label{app:imagenet1k}

\paragraph{Models and stimuli.}
We evaluated three pretrained architectures --- ResNet-50, ViT-B/16, and ConvNeXt-V2 (FCMAE-pretrained) --- on the ImageNet-1k validation set (50,000 images, $K=1000$ classes). This analysis is model-only and requires no matched human data, and is therefore independent of the GEN benchmark in images, categories, and label-space size. Because clean-data confusion matrices at $K=1000$ are prohibitively sparse (${\sim}8$ errors per class), we evaluated each model under three ImageNet-C corruption families spanning distinct corruption types (Gaussian noise, defocus blur, contrast) at severities 3 and 5, raising error density to ${\sim}20$--$44$ errors per class. Conditions were only analyzed when they reached at least 20 errors per class (11/18 conditions met this criterion and all are listed in Table~\ref{tab:imagenet1k_conditions} for completeness).

\paragraph{Concentration metric.}
Breadth and the pairwise asymmetry statistics are not estimable at this density, since the number of off-diagonal cells ($K(K-1)=999{,}000$) vastly exceeds the number of errors. We therefore report the effective number of error targets per class, $n_1 = \exp(H)$, where $H$ is the Shannon entropy of the row's off-diagonal error distribution, averaged over classes. This Hill number is well defined at low counts and is reported relative to density-matched references rather than interpreted in absolute terms.

\paragraph{Density-matched references.}
For each real confusion matrix we generated reference matrices at $K=1000$ from the same broad--weak and sink cost geometries used in the generative simulation (Section~\ref{sec_sim}), calibrated to the real matrix's per-class error count and trial count. Because the diffuse reference and the real matrix share an identical error density, finite-sample bias affects both equally and cancels in the comparison. The diffuse reference is stable across draws (46.1--46.5 effective targets), confirming that the comparison is not sensitive to the particular reference sample, but the concentration is not localizable to a subset of classes. The $50\times50$ block formed by the heaviest-receiving classes and the classes feeding them carries only ${\sim}3.5\%$ of total error mass, so no sub-block of the $1000\times1000$ matrix represents the structure and the matrix cannot be displayed directly at this scale. Concentration at $K=1000$ is therefore per-class --- each class spreads its errors over few targets --- without a shared set of sink categories, the large-label-space counterpart of the shifting-receiver result in Section~\ref{sec:results_robust}.

\begin{figure}[tbp]
  \centering
  \includegraphics[width=\linewidth]{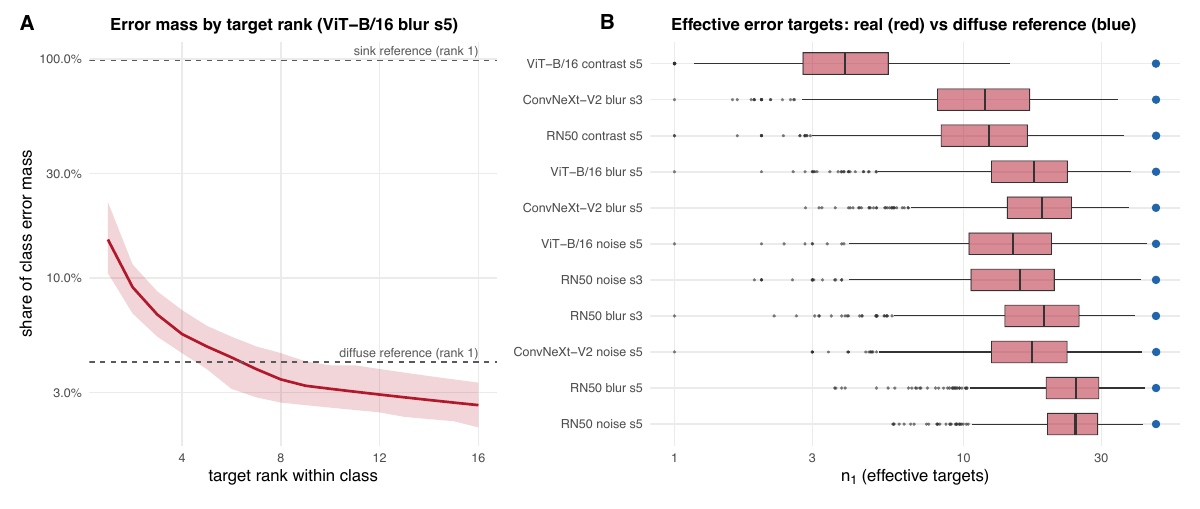}
  \caption{\textbf{Error concentration at $K=1000$ against density-matched references.} \textbf{(A)}~Share of a class's error mass falling on its $k$-th most-confused target for ViT-B/16 under defocus blur at severity~5, the condition closest to the median effective-target count across the 11 analyzed conditions (median, with interquartile range across classes), truncated at the last rank defined for at least three quarters of classes. Dashed lines mark the rank-1 share of the density-matched diffuse and sink references. The real profile sits between them, concentrated but not degenerate. \textbf{(B)}~Effective number of error targets $n_1$ per class (red) against the diffuse reference (blue), for all 11 conditions meeting the density criterion, ordered by separation from the reference with the most concentrated (ViT-B/16, contrast, severity~5) at the top. The matrices themselves are not shown: at $K=1000$ no compact sub-block carries the structure, so error structure is displayed as profiles rather than as a matrix. Reference values are those reported in Table~\ref{tab:imagenet1k_conditions} and the references are summarized rather than re-drawn here, so this figure and the reported statistics rest on identical quantities.}
  \label{fig:concentration_K1000}
\end{figure}

\begin{table}[t]
\caption{\textbf{Error concentration at $K=1000$, by condition.} The 11 of 18 model$\times$corruption$\times$severity conditions meeting the density criterion ($\geq 20$ errors per class), ordered by density. ``Eff.\ targets'' is the effective number of error targets per class, $\exp(H)$; ``diffuse'' is the density-matched broad--weak reference. Fold is the ratio of real to reference top-target share (${\sim}0.041$ in all conditions).}
\label{tab:imagenet1k_conditions}
\centering\small
\setlength{\tabcolsep}{5pt}
\begin{tabular}{llcccccc}
\toprule
& & & Errors & Top-target & \multicolumn{2}{c}{Eff.\ targets} & \\
\cmidrule(lr){6-7}
Model & Corruption & Sev. & per class & share & Real & Diffuse & Fold \\
\midrule
ResNet-50    & Defocus blur   & 5 & 43.8 & 0.147 & 24.1 & 46.4 & 3.6$\times$ \\
ResNet-50    & Gaussian noise & 5 & 42.0 & 0.134 & 24.3 & 46.3 & 3.3$\times$ \\
ConvNeXt-V2  & Defocus blur   & 5 & 36.1 & 0.183 & 18.9 & 46.3 & 4.4$\times$ \\
ResNet-50    & Defocus blur   & 3 & 33.2 & 0.173 & 19.6 & 46.3 & 4.2$\times$ \\
ViT-B/16     & Defocus blur   & 5 & 30.4 & 0.184 & 17.8 & 46.4 & 4.4$\times$ \\
ConvNeXt-V2  & Gaussian noise & 5 & 26.8 & 0.164 & 18.0 & 46.4 & 4.0$\times$ \\
ResNet-50    & Gaussian noise & 3 & 25.1 & 0.180 & 16.1 & 46.3 & 4.3$\times$ \\
ViT-B/16     & Gaussian noise & 5 & 24.1 & 0.182 & 15.7 & 46.3 & 4.4$\times$ \\
ViT-B/16     & Contrast       & 5 & 24.1 & 0.576 & 4.3  & 46.3 & 13.9$\times$ \\
ConvNeXt-V2  & Defocus blur   & 3 & 23.3 & 0.237 & 12.7 & 46.4 & 5.7$\times$ \\
ResNet-50    & Contrast       & 5 & 21.2 & 0.219 & 12.7 & 46.4 & 5.3$\times$ \\
\bottomrule
\end{tabular}
\end{table}

\paragraph{Scope.}
This experiment establishes that the concentration signature generalizes to a substantially larger label space and to transformer and self-supervised architectures. It does not establish that the full directional breadth--strength decomposition generalizes to $K=1000$, which remains under-determined at achievable error densities, nor does it extend to other task paradigms such as detection or segmentation, which do not yield a comparable $n$-way confusion matrix without further design choices.

\section{Secondary diagnostics}
\label{sec:sim_secondary_diagnostics}
\subsection{Non-monotone frontier segments}
\label{sec:rd_nonmonotone}
We summarize RD geometry from the inferred costs using two complementary notions of \emph{slope}:
\begin{enumerate}[leftmargin=*]
\item \textbf{Global RD slope and curvature (primary).}
For each cost matrix ($\rho_{\mathrm{true}}$ and $\hat\rho$) we compute an RD curve by sweeping an inverse-temperature parameter $\lambda$ over a fixed grid and recording $(R(\lambda), D(\lambda))$. We define the global slope signature $\beta$ as the median finite-difference derivative along the frontier, $\beta = \mathrm{median}\{\Delta R/\Delta D\}$. We define curvature $\kappa$ as the variance of these finite-difference slopes, and efficiency (AUC) as the trapezoidal area under the RD curve.
In 25 of the 1650 GEN blocks the Blahut--Arimoto sweep produces a numerically non-monotone frontier segment, yielding a positive $\beta$. These are near-deterministic channels --- mean row entropy $2\times10^{-4}$ and mean row-max $1.000$, against $0.93$ and $0.70$ for the analysis set as a whole --- so $\hat{\rho}$ is poorly identified over part of the range. Twenty-four are removed by the collapsed-channel filter (Appendix~\ref{app:collapse_filter}). The remaining block narrowly misses both criteria (mean row entropy $0.023$ against the $10^{-3}$ cutoff; mean row-max $0.996$ against $0.999$) and is retained in all reported analyses; excluding it changes no group-wise Spearman correlation by more than $0.004$.

\item \textbf{Local operating-point slope (secondary).}
We compute the empirical operating rate $R^\star = I(X;Y)$ from the sampled confusion \emph{counts} (using the empirical stimulus prior), then recover the local slope $s$ by root-finding for the value of $\lambda$ at which the optimal channel under cost $\rho$ attains mutual information $R^\star$.
\end{enumerate}

\subsection{Simulation Results}
\label{sec:sim_results_appx}

\paragraph{Channel Generation, Sampling, and RD Model Fitting.}
Given $\rho_{\mathrm{true}}$, a generation inverse temperature $\lambda_{\text{gen}}$, and a uniform stimulus prior $p(x)=1/K$, we generate channels via BA iterations and draw counts via
\[
N_{i\cdot} \sim \text{Multinomial}(N_{\text{per row}}, q_{\lambda_{\text{gen}}}(\cdot \mid x=i)),
\]
before recovering $\hat\rho$ by applying the same MAP RD pipeline used for empirical systems (see Section~\ref{sec:confusion_matrices}).

\paragraph{Simulation Grid and Diagnostics.}
We compute RD signatures from both $\rho_{\mathrm{true}}$ and $\hat\rho$, and compute asymmetry metrics from the corresponding induced channels/sampled confusions. Simulations are run over grids varying antisymmetry magnitude $a$, generation inverse temperature $\lambda_{\mathrm{gen}}$, and per-class trial count $N_{\text{per row}}$, across both antisymmetry structures (broad--weak, sink) with multiple seeds. We reserve $\beta$ for the empirical RD slope metric (generalization strength) and use $\lambda$ to denote inverse-temperature parameters in Blahut–Arimoto and simulation generation. For pairwise breadth/strength summaries we threshold asymmetric pairs at $\varepsilon=10^{-6}$, and for each replicate we collect RD metrics, asymmetry scores, and recovery diagnostics. We also fit replicate-level mixed-effects models and apply BH--FDR correction across trend tests (Table~\ref{tab:sim_mixed_anova_key_A5}). Recovery is validated against known ground truth by the correlation between the symmetric components of $\rho_{\mathrm{true}}$ and $\hat\rho$, which defines the recovery criterion in Table~\ref{tab:sim_recovery_A3}. Slice-wise residual regressions are in Table~\ref{tab:sim_residual_regressions_A4}.

\paragraph{Recovery sensitivity and regime-wise pass-rate tests.}
We report full recovery tables and per-slice two-sample proportion tests (BH--FDR), confirming broad--weak \(>\) sinks pass rates in all \(15\) slices and FDR significance in \(10/15\). We additionally summarize how recovery rates vary with \(\lambda_{\mathrm{gen}}\) and \(N_{\mathrm{per\,row}}\) and emphasize that strict-recovery contrasts are conditional on identifiability, not unconditional mechanism differences.

\paragraph{Component-wise recovery of $\rho$ is mechanism-dependent.}
We compared the symmetric and antisymmetric components of $\hat\rho$ against ground truth across the full grid. The antisymmetric correlation is undefined for the $300$ runs at $a=0$, where the true antisymmetric component vanishes. Among the remainder, recovery of the antisymmetric component is confined to the broad--weak generator, where the median correlation rises with antisymmetry magnitude (from $0.12$ at $a=0.1$ to $0.39$ at $a=2.0$) and with sample size (at $a=2.0$, from $0.30$ at $N_{\mathrm{per\,row}}=50$ to $0.44$ at $N_{\mathrm{per\,row}}=1000$). For the sink generator it does not recover at any setting, with medians between $-0.02$ and $-0.01$ at every $a$ and every $N_{\mathrm{per\,row}}$. The symmetric component recovers moderately and declines as antisymmetry grows ($0.60$ at $a=0$; $0.19$ broad--weak and $0.14$ sinks at $a=2.0$) without improving at larger $N_{\mathrm{per\,row}}$, indicating an identifiability rather than a sampling limit. This is the mechanism behind the pass-rate gap in Table~\ref{tab:sim_recovery_A3}: the directional component of sink-like costs is not identified by the MAP pipeline. Accordingly $\hat\rho$ should not be read as an estimate of directional cost structure for sink-like systems. The asymmetry metrics reported empirically are computed directly from the confusion matrices rather than from $\hat\rho$ (Section~\ref{sec:asym_metrics}), so they are unaffected.

\paragraph{Slope and curvature behave differently from AUC: magnitude is sufficient.}
\label{sec:sim_breadth_strength}
The residualized RD slope magnitude and curvature show a complementary pattern. For slope magnitude, global magnitude is highly predictive across all regimes (Frobenius coefficient negative in \(15/15\) slices; median \(-0.144\), range \([-0.202,\,-0.063]\), BH--FDR), whereas breadth is weaker and less reliable (median \(-0.783\), significant in \(3/15\) slices) and strength is more consistently negative (median \(-1.95\), significant in \(10/15\) slices). For curvature (log-transformed), global magnitude is again uniformly predictive (Frobenius coefficient negative in \(15/15\) slices; median \(-0.201\), range \([-0.438,\,-0.089]\), BH--FDR), while breadth/strength terms are inconsistent. Thus, AUC seems to be the RD summary for which the \emph{organization} of asymmetry (breadth vs strength) matters most, whereas slope and curvature primarily reflect overall directional magnitude.

\paragraph{Finite-sample error diagnostics for fitted versus true RD geometry.}
Recovery error is regime-dependent, where larger samples mainly stabilize the recovered AUC and reduce its dispersion, while slope-related quantities can fail sporadically in a mechanism-specific way. This motivates using non-collapsed data for primary trends and strict recovery as a conservative sensitivity analysis.

\paragraph{Saturation behavior for sink-like organization.}
Sink-like organization transitions rapidly into a sparse, high-strength regime, where breadth declines steeply and then saturates as antisymmetry grows. This supports the interpretation that sink-like directionality quickly concentrates into a small set of one-way confusions and then becomes insensitive to further increases in antisymmetry strength.

\paragraph{Strict-recovery robustness of residual effects.}
Finally, we repeat the key accuracy-residualized regressions under strict recovery and summarize sign/stability of the main effects in Table~\ref{tab:sim_sign_stability_A6}. This confirms that the qualitative pattern---AUC residuals requiring breadth--strength organization, and slope/curvature residuals tracking global magnitude---persists under conservative filtering.

\begin{table}[!tbp]
\caption{\textbf{Block-wise asymmetry summary.} One value per unique experiment$\times$condition$\times$model block. Values are mean$\pm$SE across blocks.}
\label{tab:asymmetry_appendix_summary_A1}
\standardtable

\setlength{\tabcolsep}{4pt}
\renewcommand{\arraystretch}{1.15}

\begin{tabularx}{\textwidth}{@{}Y C{0.14\textwidth} C{0.18\textwidth} C{0.18\textwidth}@{}}
\toprule
Metric & Blocks (A/H) & ANNs (mean$\pm$SE) & Humans (mean$\pm$SE) \\
\midrule
Frobenius asymmetry index
& 1569 / 81 & $1.220 \pm 0.0047$ & $1.044 \pm 0.0097$ \\
\# asymmetric pairs ($n_{\mathrm{pairs}}$), all ANNs
& 1569 / 81 & $64.16 \pm 0.666$ & $85.41 \pm 2.76$ \\
\# asymmetric pairs ($n_{\mathrm{pairs}}$), baseline ANNs
& 243 / 81 & $53.95 \pm 1.45$ & $85.41 \pm 2.76$ \\
Fraction of asymmetric pairs
& 1569 / 81 & $0.535 \pm 0.0056$ & $0.712 \pm 0.023$ \\
Conditional strength $\bar{\Delta} = \mathbb{E}_{i<j}[\lvert C_{ij}-C_{ji}\rvert \mid \lvert C_{ij}-C_{ji}\rvert>\varepsilon]$
& 1569 / 81 & $0.1409 \pm 0.0049$ & $0.0422 \pm 0.0022$ \\
\bottomrule
\end{tabularx}
\end{table}

\begin{table*}[!tbp]
\centering\footnotesize
\setlength{\tabcolsep}{3pt}
\renewcommand{\arraystretch}{1.15}
\caption{\textbf{Planned humans vs.\ group comparisons.} Frobenius asymmetry index $A_F^{\mathrm{off}}$ and accuracy, each group against humans (asymmetry $1.044$, accuracy $0.564$; $n=81$ blocks). $\Delta$ is the group mean minus the human mean, with the Welch--Satterthwaite 95\% CI. The rank column is the two-sided Wilcoxon rank-sum test, the parametric column Welch's $t$; both are Benjamini--Hochberg corrected across the six groups within each metric. Rank-based tests are primary (Section~\ref{sec:asym_metrics}) and the two agree everywhere except Distortion-trained accuracy, where the parametric test is marginal. Blocks are collapse-filtered ($n=1610$ retained); Wilcoxon $W$ statistics are in the released analysis output.}
\label{tab:group_tests_A7}
\begin{tabular}{@{}lrrrcrrrr@{}}
\toprule
& & & & & \multicolumn{2}{c}{Welch} & \multicolumn{2}{c}{$p_{\mathrm{BH}}$} \\
\cmidrule(lr){6-7}\cmidrule(lr){8-9}
Group & $n$ & Mean & $\Delta$ & 95\% CI & $t$ & df & rank & param. \\
\midrule
\multicolumn{9}{@{}l}{\emph{Frobenius asymmetry}}\\
GoogLeNet          & 81   & 1.339 & $+0.295$ & $[0.271,\,0.319]$   & 24.55   & 146.6 & $8.3{\times}10^{-27}$ & $5.2{\times}10^{-53}$ \\
ResNet-152         & 80   & 1.315 & $+0.271$ & $[0.239,\,0.302]$   & 16.92   & 148.1 & $2.7{\times}10^{-22}$ & $4.0{\times}10^{-36}$ \\
VGG-19             & 80   & 1.329 & $+0.285$ & $[0.256,\,0.314]$   & 19.42   & 156.2 & $4.4{\times}10^{-24}$ & $5.0{\times}10^{-43}$ \\
Distortion-trained & 1182 & 1.210 & $+0.166$ & $[0.144,\,0.188]$   & 14.84   & 139.0 & $9.9{\times}10^{-18}$ & $2.8{\times}10^{-30}$ \\
Specialised        & 53   & 1.022 & $-0.023$ & $[-0.065,\,0.020]$  & $-1.06$ & 79.0  & $0.136$ & $0.291$ \\
All-noise          & 53   & 1.007 & $-0.037$ & $[-0.078,\,0.004]$  & $-1.79$ & 81.6  & $0.066$ & $0.093$ \\
\addlinespace
\multicolumn{9}{@{}l}{\emph{Accuracy}}\\
GoogLeNet          & 81   & 0.512 & $-0.052$ & $[-0.157,\,0.052]$  & $-0.99$ & 157.4 & $0.923$ & $0.371$ \\
ResNet-152         & 80   & 0.516 & $-0.048$ & $[-0.154,\,0.058]$  & $-0.90$ & 155.1 & $0.923$ & $0.371$ \\
VGG-19             & 80   & 0.497 & $-0.067$ & $[-0.170,\,0.037]$  & $-1.28$ & 156.6 & $0.750$ & $0.306$ \\
Distortion-trained & 1182 & 0.482 & $-0.083$ & $[-0.155,\,-0.011]$ & $-2.28$ & 93.9  & $0.612$ & $0.050$ \\
Specialised        & 53   & 0.789 & $+0.225$ & $[0.138,\,0.311]$   & 5.14    & 131.5 & $4.8{\times}10^{-6}$ & $6.0{\times}10^{-6}$ \\
All-noise          & 53   & 0.775 & $+0.211$ & $[0.119,\,0.302]$   & 4.54    & 131.0 & $4.8{\times}10^{-6}$ & $3.8{\times}10^{-5}$ \\
\bottomrule
\end{tabular}
\end{table*}

\begin{table*}[tbp]
\appendixtablestyle
\caption{\textbf{Asymmetry--RD associations by group.}
Spearman correlations between Frobenius asymmetry and RD metrics (AUC, $\beta_{\mathrm{median}}$, $\log_{10}(\kappa{+}1)$), plus block-demeaned slope differences ($\Delta$slope) vs.\ humans. $^{*}p<0.05$, $^{**}p<0.01$, $^{***}p<0.001$.}
\label{tab:asym_rd_by_group_A2}

\textbf{(A) Rank correlations}\par\vspace{0.5ex}
\centering
\begin{tabular}{lrrrccc}
\toprule
\textbf{Group} & $n$ & \%Coll. & $\rho_{\text{AUC}}$ & $\rho_\beta$ & $\rho_\kappa$ \\
\midrule
Humans        & 81   & 0.00 &  0.29 &  0.17 & -0.22 \\
GoogLeNet     & 81   & 0.00 & -0.40 & -0.33 &  0.43 \\
ResNet-152    & 80   & 1.23 & -0.39 & -0.46 &  0.41 \\
VGG-19        & 80   & 1.23 & -0.43 & -0.35 &  0.48 \\
Distortion-trained & 1182 & 3.11 & -0.73 & -0.57 &  0.72 \\
Specialised   & 53   & 0.00 &  0.37 &  0.18 & -0.31 \\
All-noise     & 53   & 0.00 &  0.32 &  0.22 & -0.36 \\
\bottomrule
\end{tabular}

\vspace{1.2ex}

\textbf{(B) $\Delta$slope vs.\ humans}\par\vspace{0.5ex}
\centering
\begin{tabular}{lccc}
\toprule
\textbf{Group} & AUC & $\beta$ & $\kappa$ \\
\midrule
Humans        & \multicolumn{3}{c}{0 (ref.)} \\
GoogLeNet     & 2.67 ($p$=.34) & 0.91 (.29) & -0.87 (.62) \\
ResNet-152    & -0.69 (.74) & 0.36 (.58) & -0.48 (.71) \\
VGG-19        & -1.36 (.57) & -0.35 (.63) & 0.61 (.68) \\
Distortion-trained & \textbf{-7.89} ($1.3{\times}10^{-7}$)$^{***}$ & \textbf{-1.21} (.0085)$^{**}$ & \textbf{3.96} ($2{\times}10^{-5}$)$^{***}$ \\
Specialised   & \textbf{-4.77} (.014)$^{*}$ & \textbf{-1.22} (.042)$^{*}$ & \textbf{3.40} (.0049)$^{**}$ \\
All-noise     & -3.55 (.055) & -0.83 (.14) & \textbf{2.56} (.026)$^{*}$ \\
\bottomrule
\end{tabular}
\end{table*}

\begin{table}[htbp]
\centering
\caption{Recovery sensitivity. A run is counted as recovered if the correlation between the recovered and true symmetric components of $\rho$ exceeds 0.2. Fractions are computed over all 60 runs per slice and structure, including numerically collapsed runs (26 collapsed runs pass this criterion). The table reports recovery fractions within each $(\lambda_{\mathrm{gen}}, N_{\mathrm{per\,row}})$ slice and BH--FDR adjusted $p$-values for differences in recovery rates (two-sample proportion tests).}
\label{tab:sim_recovery_A3}
\begin{tabular}{cc cc c c}
\toprule
$\lambda_{\mathrm{gen}}$ & $N_{\mathrm{per\,row}}$ & 
\makecell{Frac.\ recovered \\ (broad--weak)} & 
\makecell{Frac.\ recovered \\ (sinks)} & 
$\Delta$ frac. & $p_{\mathrm{FDR}}$ \\
\midrule
0.2 &   50 & 0.567 & 0.433 & 0.133 & 0.232 \\
0.2 &  200 & 0.700 & 0.583 & 0.117 & 0.271 \\
0.2 & 1000 & 0.867 & 0.633 & 0.233 & 0.009 \\
0.5 &   50 & 0.667 & 0.517 & 0.150 & 0.172 \\
0.5 &  200 & \textbf{0.867} & \textbf{0.450} & \textbf{0.417} & $<$0.001 \\
0.5 & 1000 & 0.817 & 0.483 & 0.333 & $<$0.001 \\
1.0 &   50 & 0.817 & 0.500 & 0.317 & 0.001 \\
1.0 &  200 & 0.783 & 0.433 & 0.350 & $<$0.001 \\
1.0 & 1000 & 0.817 & 0.517 & 0.300 & 0.002 \\
2.0 &   50 & 0.817 & 0.533 & 0.283 & 0.003 \\
2.0 &  200 & 0.817 & 0.417 & 0.400 & $<$0.001 \\
2.0 & 1000 & 0.767 & 0.400 & 0.367 & $<$0.001 \\
5.0 &   50 & 0.900 & 0.517 & 0.383 & $<$0.001 \\
5.0 &  200 & 0.833 & 0.750 & 0.083 & 0.369 \\
5.0 & 1000 & 0.817 & 0.633 & 0.183 & 0.056 \\
\bottomrule
\end{tabular}
\end{table}

\begin{table*}[tbp]
\centering
\setlength{\tabcolsep}{4pt}
\renewcommand{\arraystretch}{1.15}
\caption{\textbf{Residual regression summary.}
Within each $\left(\lambda_{\mathrm{gen}}, N_{\mathrm{per\,row}}\right)$ slice, we first residualize each RD signature by mean diagonal probability (accuracy proxy), then regress the residual on either (i) breadth/strength terms or (ii) a global asymmetry magnitude term. ``Structure offset'' refers to contrast between sink-like and broad--weak configurations. Coefficients are summarized across slices; significance counts use BH--FDR within-slice tests.}
\label{tab:sim_residual_regressions_A4}

\begin{tabularx}{\textwidth}{@{}l l Y r r r c c@{}}
\toprule
Outcome & Predictor set & Term & Median & Min & Max & \shortstack{Sig.\\slices} & Slices \\
\midrule
AUC & components & Breadth & -2.05 & -5.40 &  1.52 & 11 & 15 \\
AUC & components & Strength & -4.28 & -11.56 & 2.45 & 12 & 15 \\
AUC & components & Breadth $\times$ Strength &  5.33 & -4.65 & 16.27 & 5 & 15 \\
AUC & components & Structure offset & -0.447 & -0.657 & -0.162 & 15 & 15 \\
AUC & magnitude & Global asymmetry mag. & -0.0047 & -0.0470 & 0.206 & 3 & 15 \\
AUC & magnitude & Structure offset & -0.482 & -0.691 & -0.171 & 15 & 15 \\
\addlinespace
Global slope & components & Breadth & -0.78 & -2.12 & 0.37 & 3 & 15 \\
Global slope & components & Strength & -1.95 & -3.70 & -0.40 & 10 & 15 \\
Global slope & components & Breadth $\times$ Strength & -5.02 & -9.96 & 5.73 & 7 & 15 \\
Global slope & components & Structure offset & -0.188 & -0.357 & 0.066 & 9 & 15 \\
Global slope & magnitude & Global asymmetry mag. & -0.144 & -0.202 & -0.063 & 15 & 15 \\
Global slope & magnitude & Structure offset & -0.243 & -0.386 & -0.081 & 15 & 15 \\
\addlinespace
Curvature & components & Breadth & -0.048 & -0.167 & 0.129 & 3 & 15 \\
Curvature & components & Strength & -0.165 & -0.540 & 0.084 & 6 & 15 \\
Curvature & components & Breadth $\times$ Strength &  0.066 & -1.030 & 1.250 & 2 & 15 \\
Curvature & components & Structure offset &  0.058 & -0.011 & 0.124 & 7 & 15 \\
Curvature & magnitude & Global asymmetry mag. & -0.201 & -0.438 & -0.089 & 15 & 15 \\
Curvature & magnitude & Structure offset &  0.060 & -0.009 & 0.126 & 8 & 15 \\
\bottomrule
\end{tabularx}
\end{table*}

\begin{table}[!tbp]
\centering
\caption{\textbf{Mixed-effects ANOVA key terms.} Selected fixed-effect tests from replicate-level mixed models $Y\sim\texttt{structure}\times a\times\lambda_{\mathrm{gen}}\times\log_{10}N_{\mathrm{per\,row}}+(1|\texttt{cell})$.}
\label{tab:sim_mixed_anova_key_A5}
\standardtable

\begin{tabular}{llll}
\toprule
Outcome & Term & $F$ & $p$ \\
\midrule
auc\_true & structure & 10.16 & 0.00146 \\
auc\_true & a\_scale & 1176 & 2.2e-16 \\
auc\_true & structure:a\_scale & 877.1 & 2.2e-16 \\
beta\_median\_true\_pos & structure & 24.41 & 7.8e-07 \\
beta\_median\_true\_pos & a\_scale & 827.7 & 2.2e-16 \\
beta\_median\_true\_pos & structure:a\_scale & 613.4 & 2.2e-16 \\
kappa\_true & structure & 5.34 & 0.0209 \\
kappa\_true & a\_scale & 964.7 & 2.2e-16 \\
kappa\_true & structure:a\_scale & 701.8 & 2.2e-16 \\
\bottomrule
\end{tabular}
\end{table} 

\begin{table}[!tbp]
\caption{\textbf{Strict-recovery robustness of residual AUC effects.} For each $(\lambda_{\mathrm{gen}}, N_{\mathrm{per\,row}})$ slice and structure, we regress accuracy-residualized AUC on antisymmetry strength $a$ under the no-collapse filter (all non-collapsed runs) and the strict-recovery filter (additionally requiring symmetric-component recovery correlation $>0.2$).
Columns show the slope sign and BH--FDR adjusted $p$-value under each filter. Signs agree in all 30 rows --- broad--weak positive in 15/15 slices and sinks negative in 15/15 under both filters --- so the qualitative pattern is robust to identifiability constraints. However, the two structures differ in what strict recovery costs them: broad--weak slopes remain at $p<0.001$ in every slice, whereas sink slopes fall from $p<0.001$ in 11/15 slices to 4/15, consistent with the recovery gap in Table~\ref{tab:sim_recovery_A3}.}
\label{tab:sim_sign_stability_A6}
\centering
\small
\setlength{\tabcolsep}{4pt}
\renewcommand{\arraystretch}{1.1}
\begin{tabular}{cc l rr rr}
\toprule
& & & \multicolumn{2}{c}{No-collapse} & \multicolumn{2}{c}{Strict-recovery} \\
\cmidrule(lr){4-5}\cmidrule(lr){6-7}
$\lambda_{\mathrm{gen}}$ & $N_{\mathrm{per\,row}}$ & Structure &
Sign & $p_{\mathrm{FDR}}$ & Sign & $p_{\mathrm{FDR}}$ \\
\midrule
0.2 &   50 & Broad--weak & $+$ & $<$0.001 & $+$ & $<$0.001 \\
0.2 &   50 & Sinks       & $-$ & $<$0.001 & $-$ & 0.003     \\
0.2 &  200 & Broad--weak & $+$ & $<$0.001 & $+$ & $<$0.001 \\
0.2 &  200 & Sinks       & $-$ & $<$0.001 & $-$ & $<$0.001 \\
0.2 & 1000 & Broad--weak & $+$ & $<$0.001 & $+$ & $<$0.001 \\
0.2 & 1000 & Sinks       & $-$ & 0.002     & $-$ & 0.002     \\
0.5 &   50 & Broad--weak & $+$ & $<$0.001 & $+$ & $<$0.001 \\
0.5 &   50 & Sinks       & $-$ & $<$0.001 & $-$ & 0.019     \\
0.5 &  200 & Broad--weak & $+$ & $<$0.001 & $+$ & $<$0.001 \\
0.5 &  200 & Sinks       & $-$ & $<$0.001 & $-$ & 0.042     \\
0.5 & 1000 & Broad--weak & $+$ & $<$0.001 & $+$ & $<$0.001 \\
0.5 & 1000 & Sinks       & $-$ & $<$0.001 & $-$ & 0.029     \\
1.0 &   50 & Broad--weak & $+$ & $<$0.001 & $+$ & $<$0.001 \\
1.0 &   50 & Sinks       & $-$ & $<$0.001 & $-$ & $<$0.001 \\
1.0 &  200 & Broad--weak & $+$ & $<$0.001 & $+$ & $<$0.001 \\
1.0 &  200 & Sinks       & $-$ & $<$0.001 & $-$ & 0.006     \\
1.0 & 1000 & Broad--weak & $+$ & $<$0.001 & $+$ & $<$0.001 \\
1.0 & 1000 & Sinks       & $-$ & $<$0.001 & $-$ & $<$0.001 \\
2.0 &   50 & Broad--weak & $+$ & $<$0.001 & $+$ & $<$0.001 \\
2.0 &   50 & Sinks       & $-$ & $<$0.001 & $-$ & $<$0.001 \\
2.0 &  200 & Broad--weak & $+$ & $<$0.001 & $+$ & $<$0.001 \\
2.0 &  200 & Sinks       & $-$ & $<$0.001 & $-$ & 0.049     \\
2.0 & 1000 & Broad--weak & $+$ & $<$0.001 & $+$ & $<$0.001 \\
2.0 & 1000 & Sinks       & $-$ & $<$0.001 & $-$ & 0.012     \\
5.0 &   50 & Broad--weak & $+$ & $<$0.001 & $+$ & $<$0.001 \\
5.0 &   50 & Sinks       & $-$ & 0.005     & $-$ & 0.012     \\
5.0 &  200 & Broad--weak & $+$ & $<$0.001 & $+$ & $<$0.001 \\
5.0 &  200 & Sinks       & $-$ & 0.004     & $-$ & 0.011     \\
5.0 & 1000 & Broad--weak & $+$ & $<$0.001 & $+$ & $<$0.001 \\
5.0 & 1000 & Sinks       & $-$ & 0.012     & $-$ & 0.013     \\
\bottomrule
\end{tabular}
\end{table}


\clearpage
\section*{NeurIPS Paper Checklist} 


\begin{enumerate}

\item {\bf Claims}
    \item[] Question: Do the main claims made in the abstract and introduction accurately reflect the paper's contributions and scope?
    \item[] Answer: \answerYes{}.
    \item[] Justification: The abstract and introduction state all four main findings. The human--model comparison is bounded to $K=16$ visual categorization under controlled perturbation, while the model-side generalization to $K=1000$ is stated with its own scope limits in Appendix~\ref{app:imagenet1k}.

\item {\bf Limitations}
    \item[] Question: Does the paper discuss the limitations of the work performed by the authors?
    \item[] Answer: \answerYes{} 
    \item[] Justification: The dedicated Limitations paragraph in the Discussion addresses label granularity, human sample size, MAP prior sensitivity, mechanism-dependent recoverability, and the representational interpretation.

\item {\bf Theory assumptions and proofs}
    \item[] Question: For each theoretical result, does the paper provide the full set of assumptions and a complete (and correct) proof?
    \item[] Answer: \answerNA{} 
    \item[] Justification: The paper is empirical and simulation-based and does not contain any formal theorems or proofs.

    \item {\bf Experimental result reproducibility}
    \item[] Question: Does the paper fully disclose all the information needed to reproduce the main experimental results of the paper to the extent that it affects the main claims and/or conclusions of the paper (regardless of whether the code and data are provided or not)?
    \item[] Answer: \answerYes{} 
    \item[] Justification: The full pipeline is described in Section~\ref{sec:rd_framework} and Section 3; the MAP prior, $\lambda$ grid and signature definitions are specified in Appendix~\ref{app:map_prior}; the simulation grid is specified in Appendix~\ref{sec:sim_results_appx}; all details of the GEN and ModelZoo benchmarks are publicly available.

\item {\bf Open access to data and code}
    \item[] Question: Does the paper provide open access to the data and code, with sufficient instructions to faithfully reproduce the main experimental results, as described in supplemental material?
    \item[] Answer: \answerYes{} 
    \item[] Justification: All analysis code is available at \url{https://github.com/leylaroksan/rdt_asymmetry}, with scripts reproducing every figure and table. The derived data needed to reproduce the results without re-running MAP inference are deposited at figshare under CC BY 4.0: \url{https://doi.org/10.6084/m9.figshare.34059993}. The underlying behavioral and model data are publicly available via the GEN and ModelZoo repositories~\citep{geirhos2018generalisation,geirhos2021partial}. See Appendix~\ref{app:compute_licenses}.

\item {\bf Experimental setting/details}
    \item[] Question: Does the paper specify all the training and test details (e.g., data splits, hyperparameters, how they were chosen, type of optimizer) necessary to understand the results?
    \item[] Answer: \answerYes{} 
    \item[] Justification: All perturbation families, model architectures, training regimes, and the block-wise aggregation procedure are described in Sections 3.1--3.5 and Appendix~\ref{sec:architectures_appx}.

\item {\bf Experiment statistical significance}
    \item[] Question: Does the paper report error bars suitably and correctly defined or other appropriate information about the statistical significance of the experiments?
    \item[] Answer: \answerYes{} 
    \item[] Justification: Group comparisons use Wilcoxon rank-sum tests with BH--FDR correction, reported alongside Welch's $t$-tests and 95\% confidence intervals on the mean difference via the Welch--Satterthwaite approximation (Table~\ref{tab:group_tests_A7}). Error bars are s.e.m.\ as stated in figure captions. Accuracy-controlled regressions report slopes with standard errors, $t$-statistics and BH-corrected $p$-values under both OLS and cluster-robust (CR2) inference with Satterthwaite degrees of freedom (Appendix~\ref{app:inference}).

\item {\bf Experiments compute resources}
    \item[] Question: For each experiment, does the paper provide sufficient information on the computer resources (type of compute workers, memory, time of execution) needed to reproduce the experiments?
    \item[] Answer: \answerYes{} 
    \item[] Justification: Compute requirements are detailed in Appendix~\ref{app:compute_licenses}. The rate--distortion inference, asymmetry metrics and simulation analyses are CPU-based and were run on the Minerva HPC cluster. The ImageNet-1k evaluation additionally required GPU inference over the ImageNet validation set (Appendix~\ref{app:imagenet1k}).
    
\item {\bf Code of ethics}
    \item[] Question: Does the research conducted in the paper conform, in every respect, with the NeurIPS Code of Ethics \url{https://neurips.cc/public/EthicsGuidelines}?
    \item[] Answer: \answerYes{} 
    \item[] Justification: The research uses publicly available datasets with no human subjects recruited by the authors. There are no privacy or fairness concerns that arise.

\item {\bf Broader impacts}
    \item[] Question: Does the paper discuss both potential positive societal impacts and negative societal impacts of the work performed?
    \item[] Answer: \answerYes{} 
    \item[] Justification: This is foundational research on evaluation methodology. No direct path to harmful applications is identified, however, the work could improve alignment evaluation, which is broadly positive.
    
\item {\bf Safeguards}
    \item[] Question: Does the paper describe safeguards that have been put in place for responsible release of data or models that have a high risk for misuse (e.g., pre-trained language models, image generators, or scraped datasets)?
    \item[] Answer: \answerNA{} 
    \item[] Justification: No new models, datasets, or high-risk assets are released.

\item {\bf Licenses for existing assets}
    \item[] Question: Are the creators or original owners of assets (e.g., code, data, models), used in the paper, properly credited and are the license and terms of use explicitly mentioned and properly respected?
    \item[] Answer: \answerYes{} 
    \item[] Justification: The GEN benchmark is used under CC BY 4.0 and the ModelZoo benchmark code under the MIT License, both cited with their original papers. Pre-trained weights accessed through ModelZoo remain under their own licenses, including CC BY-NC 4.0 for SWSL ResNeXt-101. The ImageNet-1k evaluation additionally uses the ImageNet validation set under its terms of access, the ImageNet-C corruption benchmark under CC BY 4.0, and pretrained weights obtained through \texttt{timm} under the Apache License 2.0. Licenses for each are stated in Appendix~\ref{app:compute_licenses}.

\item {\bf New assets}
    \item[] Question: Are new assets introduced in the paper well documented and is the documentation provided alongside the assets?
    \item[] Answer: \answerYes{} 
    \item[] Justification: No new models or human data are introduced. The derived data supporting the analyses --- fitted distortion matrices, RD signature summaries, collapse flags, simulation grid outputs, and the $K=1000$ confusion matrices --- are deposited at \url{https://doi.org/10.6084/m9.figshare.34059993} under CC BY 4.0. Provenance and licensing of the source data are given in Appendix~\ref{app:compute_licenses}.

\item {\bf Crowdsourcing and research with human subjects}
    \item[] Question: For crowdsourcing experiments and research with human subjects, does the paper include the full text of instructions given to participants and screenshots, if applicable, as well as details about compensation (if any)? 
    \item[] Answer: \answerNA{} 
    \item[] Justification: No new human subjects were recruited. The human psychophysics data is from published benchmark repositories.

\item {\bf Institutional review board (IRB) approvals or equivalent for research with human subjects}
    \item[] Question: Does the paper describe potential risks incurred by study participants, whether such risks were disclosed to the subjects, and whether Institutional Review Board (IRB) approvals (or an equivalent approval/review based on the requirements of your country or institution) were obtained?
    \item[] Answer: \answerNA{} 
    \item[] Justification: No new human subjects research was conducted. Only existing data from published repositories was used.

\item {\bf Declaration of LLM usage}
    \item[] Question: Does the paper describe the usage of LLMs if it is an important, original, or non-standard component of the core methods in this research? Note that if the LLM is used only for writing, editing, or formatting purposes and does \emph{not} impact the core methodology, scientific rigor, or originality of the research, declaration is not required.
    \item[] Answer: \answerNA{} 
    \item[] Justification: LLMs were not used as a component of the core methodology.

\end{enumerate}

\end{document}